\documentclass[letterpaper]{article} 
\usepackage{aaai25}  
\usepackage{times}  
\usepackage{helvet}  
\usepackage{courier}  
\usepackage[hyphens]{url}  
\usepackage{graphicx} 
\usepackage{natbib}  
\usepackage{caption} 
\usepackage{algorithm}
\usepackage{algorithmic}
\usepackage{newfloat}
\usepackage{listings}
\DeclareCaptionStyle{ruled}{labelfont=normalfont,labelsep=colon,strut=off} 
\floatstyle{ruled}
\newfloat{listing}{tb}{lst}{}
\floatname{listing}{Listing}
\usepackage{subcaption}
\usepackage{makecell}
\usepackage{listings}
\usepackage{framed}
\usepackage{xcolor}
\usepackage{booktabs} 
\usepackage{lipsum}
\usepackage{enumitem}
\usepackage{xspace}
\usepackage{graphicx}
\usepackage{float}
\usepackage{caption}
\usepackage{multirow}
\usepackage{multicol}
\usepackage{cancel}
\usepackage{pifont}
\definecolor{citecolor}{HTML}{0071bc}
\definecolor{ourscolor}{HTML}{c2d1e5}
\usepackage{colortbl}  
\makeatletter
\DeclareRobustCommand\onedot{\futurelet\@let@token\@onedot}
\def\@onedot{\ifx\@let@token.\else.\null\fi\xspace}
\def\eg{\emph{e.g}\onedot} 

\def\ie{\emph{i.e}\onedot}

\makeatother

\lstdefinelanguage{json}{
	basicstyle=\small\ttfamily,
	commentstyle=\color{gray},
	stringstyle=\color{blue},
	morestring=[b]",
	morecomment=[l]{//},
	morecomment=[s]{/*}{*/},
}

\newcommand{\cmark}{\ding{51}}%
\newcommand{\xmark}{\ding{55}}%
\newcommand{\cxmark}{\bcancel{\cmark}}%

\title{AutoMMLab: Automatically Generating Deployable Models from Language Instructions for Computer Vision Tasks}
\author {
	Zekang Yang\textsuperscript{\rm1},
	Wang Zeng\textsuperscript{\rm1},
	Sheng Jin\textsuperscript{\rm2,\rm1}\thanks{Corresponding Author},
	Chen Qian\textsuperscript{\rm1},
	Ping Luo\textsuperscript{\rm2},
	Wentao Liu\textsuperscript{\rm1$\ast$}
}
\affiliations {
	\textsuperscript{\rm 1}SenseTime Research and Tetras.AI \\
	\textsuperscript{\rm 2}The University of Hong Kong\\
	\{yangzekang, zengwang, jinsheng, qianchen, liuwentao\}@tetras.ai  \quad pluo@cs.hku.hk
}

\begin{document}
	
	\maketitle
	
	\begin{abstract}
		Automated machine learning (AutoML) is a collection of techniques designed to automate the machine learning development process. While traditional AutoML approaches have been successfully applied in several critical steps of model development (\eg hyperparameter optimization), there lacks a AutoML system that automates the entire end-to-end model production workflow for computer vision.
		To fill this blank, we propose a novel request-to-model task, which involves understanding the user's natural language request and execute the entire workflow to output production-ready models. This empowers non-expert individuals to easily build task-specific models via a user-friendly language interface. To facilitate development and evaluation, we develop a new experimental platform called AutoMMLab and a new benchmark called LAMP for studying key components in the end-to-end request-to-model pipeline. 
		Hyperparameter optimization (HPO) is one of the most important components for AutoML. Traditional approaches mostly rely on trial-and-error, leading to inefficient parameter search. To solve this problem, we propose a novel LLM-based HPO algorithm, called HPO-LLaMA. Equipped with extensive knowledge and experience in model hyperparameter tuning, HPO-LLaMA achieves significant improvement of HPO efficiency.
		Dataset and code are available at https://github.com/yang-ze-kang/AutoMMLab.
	\end{abstract}
	
	\section{Introduction}
	\label{sec:intro}
	\begin{figure}[h]
		\centering
		\includegraphics[width=0.47\textwidth]{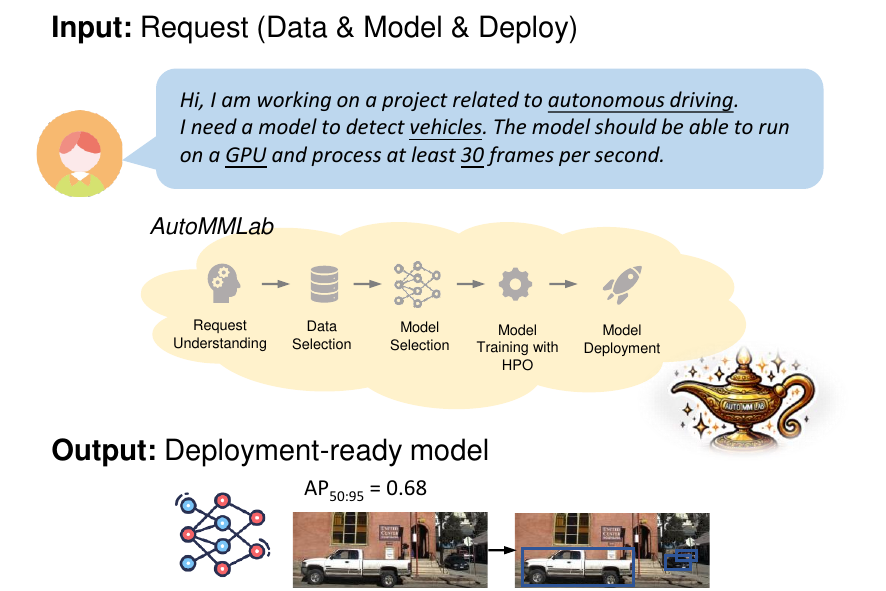}
		\caption{AutoMMLab autmatically creates deployable models from user's language instructions.}\label{fig:teaser}
	\end{figure}
	Machine learning (ML) has achieved considerable successes in recent years and an ever-growing number of disciplines rely on it. However, this success crucially relies on human ML experts to perform tasks including data preparation, model algorithm selection, model training with hyperparameter tuning and model deployment. 
	The rapid growth of machine learning applications has created a demand for off-the-shelf machine learning methods that can be used easily and without expert knowledge. 
	To this end, automated machine learning, or AutoML, has been developed in literature. AutoML aims to minimize the workload of human experts in the ML development process, making AI algorithms available and easily accessible even for non-AI experts.
	
	Recent public computer vision (CV) communities (\eg OpenMMLab) have provided a wide variety of toolboxes in various research areas such as image classification (MMPretrain~\cite{mmpretrain}), object detection (MMDetection~\cite{mmdetection}), semantic segmentation (MMSegmentation~\cite{mmseg}), and pose estimation (MMPose~\cite{mmpose}). 
	Considering large language models (LLMs) have exhibited exceptional ability in language understanding, generation, interaction, and reasoning, we are inspired to employ LLMs to connect AutoML with public CV communities (\eg OpenMMLab) for automating the \textit{whole} CV model production workflow via a natural language based interface.
	
	In this paper, we introduce a novel comprehensive task called request-to-model. 
	Unlike traditional AutoML task that focus on one particular step in the model production workflow (\eg neural architecture search or hyperparameter optimization), request-to-model extends the reach of AutoML to evolve towards fully automated model production through a user-friendly language interface. 
	This empowers individuals to leverage the capabilities of AI without extensive expertise, driving progress in various fields.
	
	To facilitate the development of the request-to-model task, we propose a LLM-empowered platform called AutoMMLab, which can serve as a testbed for developing and evaluating the request-to-model algorithms. 
	As shown in Figure~\ref{fig:teaser}, given language-based user requirements, our AutoMMLab platform will schedule each module to execute the entire workflow and finally output production-ready models for computer vision tasks. 
	AutoMMLab connects diverse datasets, CV models, training pipelines and deployment tools to facilitate the solution of numerous CV tasks. 
	The whole workflow consists of 5 major modules, including request understanding, data construction, model construction, model training with hyperparameter optimization, and model deployment. 
	
	Thanks to the modular design of AutoMMLab, we also develop the LAMP (Language-instructed Automated Model Production) benchmark for research on the language-instruction based model production. 
	The LAMP is the \emph{first} benchmark for evaluating the request-to-model AutoML algorithms in computer vision, which enables the community to explore key components in the end-to-end request-to-model pipeline. 
	
	Hyperparameter optimization (HPO) aims to choose a set of optimal hyperparameters that yields an optimal performance based on the predefined objective function. Thanks to its wide applications, it becomes one of the most important components for AutoML. 
	Traditional approaches (\eg bayesian optimization) mostly rely on trial-and-error, leading to inefficient parameter search. 
	To solve this problem, we propose a novel LLM-based hyperparameter optimization algorithm, called HPO-LLaMA. 
	Equipped with rich knowledge of model training, HPO-LLaMA can skillfully search for the optimal model training hyperparameters with significantly reduced numbers of trials.
	
	Our main contributions can be summarized as follows:
	\begin{itemize}[itemsep=2pt,topsep=0pt,parsep=0pt]
		\item We propose the novel request-to-model task, automating the whole model production workflow.
		We present the \emph{first} request-to-model AutoML platform for computer vision tasks, called AutoMMLab. By integrating AutoML and language interface, it enables non-expert users to easily build and deploy CV models, unlocking the potential of AI for a wider audience.
		\item Based on the AutoMMLab platform, we build a benchmark termed LAMP for evaluating end-to-end prompt-based model production, and also studying each component in the whole production pipeline. 
		\item We propose a novel LLM-based hyperparameter optimization algorithm, called HPO-LLaMA. To the best of our knowledge, HPO-LLaMA is the first supervised fine-tuned LLM specifically designed for HPO.
	\end{itemize}
	
	\section{Related Works}
	\textbf{Hyperparameter Optimization (HPO).}
	Grid search and random search~\cite{bergstra2012random} are commonly employed methods for hyperparameter optimization (HPO). Grid search divides the search space into regular intervals and evaluates all choices to select the best-performing one, while random search selects the best choice from a set of randomly sampled hyperparameters. 
	Bayesian optimization~\cite{smac3} is another efficient HPO method, where a surrogate model such as random forest and Gaussian process is constructed to learn from past function evaluations and choose promising future candidates.
	Gradient-based optimization methods~\cite{pedregosa2016hyperparameter,franceschi2017forward} further improve it by employing gradient information. In contrast to these traditional approaches that rely solely on trial-and-error, our proposed HPO-LLaMA leverages a large language model with extensive knowledge and experience in model hyperparameter tuning, resulting in significant improvement of HPO efficiency.
	
	\textbf{AutoML libraries and systems.}
	While several AutoML pipeline libraries and systems have been introduced, most of them specialize in specific components of the pipeline and are mainly designed for AI developers. 
	Early AutoML frameworks like Auto-WEAK~\cite{thornton2013auto} and Auto-Sklearn~\cite{feurer2015efficient} concentrated on optimizing traditional machine learning pipelines and their associated hyperparameters. More recent frameworks like Auto-PyTorch~\cite{zimmer2021auto} and AutoKeras~\cite{jin2019auto} have shifted their focus towards searching for deep learning models. 
	There are also comprehensive AutoML solution. For example, Microsoft's NNI\footnote{https://github.com/microsoft/nni} toolkit automates feature engineering, neural architecture search, hyperparameter tuning, and model compression specifically for deep learning. 
	Vega~\cite{wang2020vega} provides an extensive pipeline that encompasses data augmentation, HPO, NAS, model compression, and full training. 
	Google's AutoML\footnote{https://cloud.google.com/automl} suite offers a user-friendly collection of AutoML tools that streamline the entire machine learning pipeline.
	However, they still require intervention and effort from users with expertise.
	
	\textbf{LLM in AutoML.} Large Language Models (LLMs) are large-scale neural networks with billions of parameters that are pre-trained on massive amounts of data to achieve general-purpose language understanding and generation. 
	LLMs such as GPTs~\cite{brown2020language,openai2023gpt4}, PaLM~\cite{chowdhery2022palm} and LLaMAs~\cite{touvron2023llama,touvron2023llama2}, have demonstrated impressive capabilities in comprehension and generation of natural language.
	Some studies directly explore GPT's capabilities on AutoML including feature engineering~\cite{hollmann2023caafe}, neural architecture search (NAS)~\cite{zheng2023can}.
	For example, GENIUS~\cite{zheng2023can} proposes to use GPT-4 as a ``black box" optimiser to tackle the problem of NAS through an iterative refinement process, ~\cite{zhang2023using} demonstrates that in hyperparameter optimization tasks, LLM-based methods outperform random search and Bayesian optimization. 
	More recently, \cite{huang2024mlagentbench} and \cite{guo2024ds} leverage LLMs to iteratively optimize the baseline method for a specific task, resulting in a better-performing method.
	\cite{hong2024data} leverages LLMs to automatically analyze structured data but lacks a self-improvement strategy to refine the results.
	\cite{viswanathan2023prompt2model} presents a demo that provides a natural language task description to LLMs, and uses it to train a special-purpose NLP model that is conducive to deployment.
	Different from the work mentioned above, Our propose AutoMMLab system has several distinct features.
	First, it automates the whole model production pipeline, from understanding requirements to deploying the model, while supporting a wide range of mainstream computer vision tasks, including classification, detection, segmentation, and pose estimation.
	Second, instead of relying solely on prompt engineering, we developed a requirements understanding dataset and a hyperparameter optimization dataset, leveraging supervised fine-tuning of LLMs to enhance their domain-specific capabilities. 
	Third, it can recommend optimal hyperparameters tailored to specific requests.
	\begin{figure*}[!ht]
		\centering
		\includegraphics[width=0.9\textwidth]{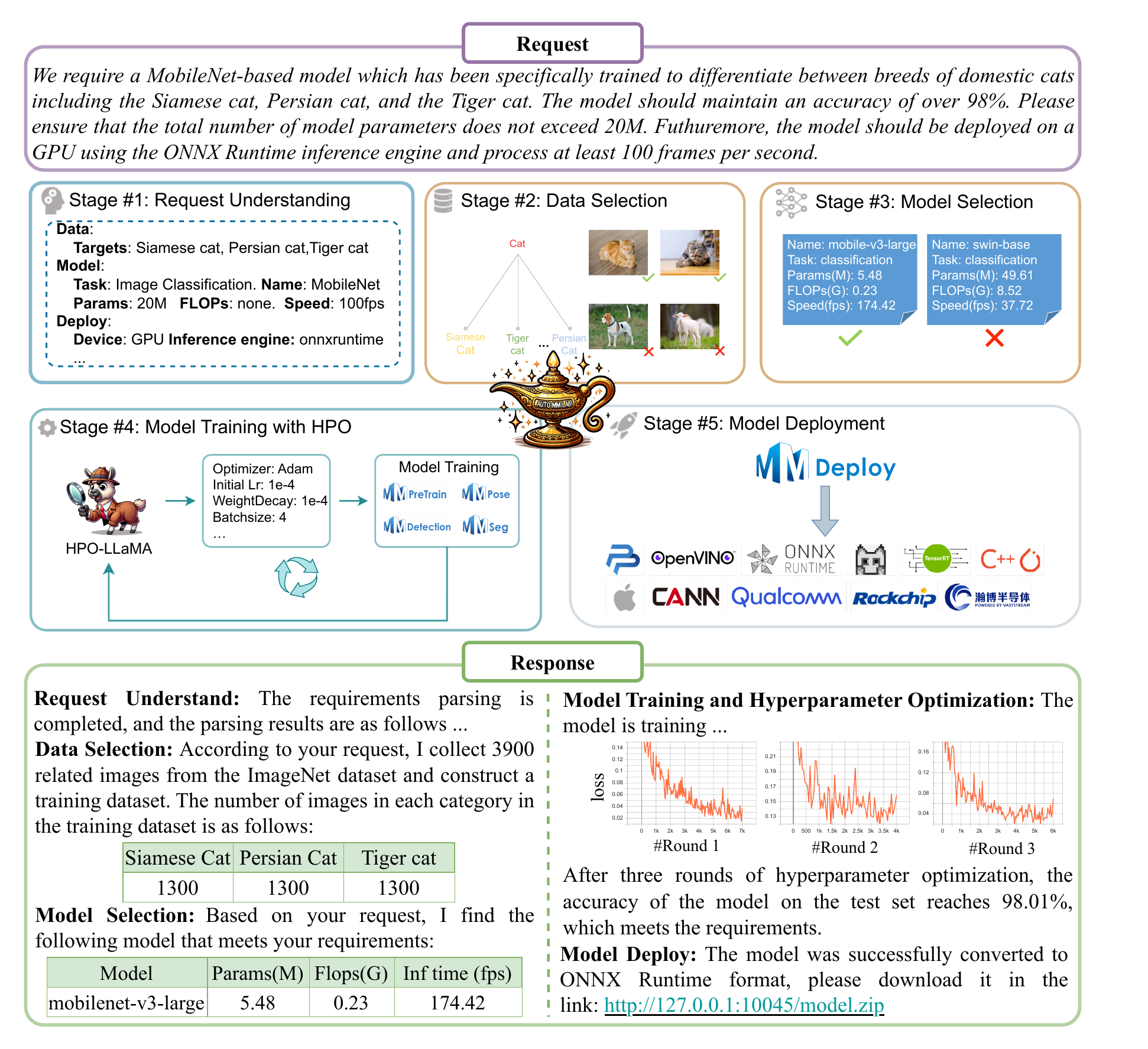}
		\caption{Overview of AutoMMLab. The workflow of AutoMMLab consists of five stages.
			\textbf{Request understanding:} Parse the language requests into formated configuration. 
			\textbf{Data selection:} Select appropriate training data from the dataset zoo.
			\textbf{Model selection:} Select the optimal model from the model zoo.
			\textbf{Model training with HPO:} Train the model and optimize the hyperparameters.
			\textbf{Model deployment:} Convert the model into a package compatible with the deployment environments.
		} 
		\label{fig:overview}
	\end{figure*}
	
	\section{AutoMMLab}
	\subsection{Overview}
	The overview of AutoMMLab platform is illustrated in Figure~\ref{fig:overview}. 
	It directly takes language-based model request from users as the input, and automatically schedule and execute the entire workflow to produce deployable models. 
	The whole pipeline of AutoMMLab consists of 5 main stages, including request understanding, data selection, model selection, model training with hyperparameter optimization (HPO), and model deployment. 
	The Request Understanding stage understands the language-based model request and schedules the model production workflow, generating a structured configuration. 
	The configuration contains detailed data information, model constraints and deployment requirements. AutoMMLab then construct the training data inside built-in dataset zoo and select the most appropriate model that meets the model constraints.
	When the data and model are ready, AutoMMLab execute model training with hyperparameter optimization (HPO) to obtain the final trianed model. 
	In the model deployment stage, AutoMMLab apply MMDeploy~\cite{mmdeploy} to deploy the model based on the user's hardware requirements.
	
	\subsection{Request Understanding} 
	\label{sec:request_understanding}
	Request understanding aims to automatically and accurately understand the user's request, schedule the model production workflow, and generate the JSON-format configuration. 
	Since it is non-trival for general-purpose LLMs to succeed in request understanding solely through prompting, we also train a task-specific LLM (RU-LLaMA) to achieve this goal. 
	
	\textbf{GPT-assisted Training Data Generation.}
	Tuning LLMs typically require numerous training data~\cite{liu2023llava,liu2023improvedllava}, which is exceptionally costly and time-consuming to collect with human crowd-scouring. Inspired by the success of recent LLMs in data generation~\cite{wang2022self}, we leverage GPT-4 to build our data generation pipeline. 
	For each task type,
	(\ie image classification, object detection, semantic segmentation, and keypoint estimation), 
	we first ask 5 professional computer vision researchers to manually design 100 diverse request-config pairs. These human annotations are used as seeds in in-context-learning to query GPT.
	In our pipeline, we first design prompts to ask GPT to generate diverse requests, and then use carefully designed prompts to query GPT to parse the requests and generate corresponding configurations.
	In our preliminary experiments, we ablated the use of GPT-3.5 and GPT-4 and empirically found that GPT-4 can consistently produce higher quality data. Therefore, we use GPT-4 in our data generation pipeline and generate 1,000 request-config pairs. Finally, the generated data samples are carefully checked and manually corrected by a group of professional annotators.
	
	\textbf{RU-LLaMA.}
	LLM has learned massive knowledge in the pre-training stage and have strong general question and answer capabilities.
	However, its usually require supervised fine-tuning on instruction-tuning dataset to better activate its domain-specific capabilities.
	We construct a instruction fine-tuning dataset based on the dataset generated by GPT-4. Following the mainstream methods of fine-tuning large language models in specific domains, we use LoRA~\cite{hu2022lora} technology to fine-tune the LLM. 
	Specifically, we utilize LoRA to fine-tune Llama-2~\cite{touvron2023llama2}, and obtained RU-LLaMA that is skilled in parsing user's requirements into JSON-format configurations.
	
	\subsection{Data Selection}
	\label{sec:data_selection}
	Data selection stage aims to use the JSON-format configuration parsed by RU-LLaMA to automatically retrieve and build request-related training dataset from the dataset zoo.
	\textbf{Dataset Zoo.}
	To facilitate dataset preparation, we construct a comprehensive dataset zoo encompassing a wide variety of public datasets~\cite{krizhevsky2017imagenet, lin2014microsoft, cordts2016cityscapes}.
	Each dataset is complemented by a data card that contains the meta information appertaining to the dataset intricately.
	We also take into account AI safety considerations, by ensuring openness, transparency, and reliability of data.
	\textbf{Dataset Selection Pipeline.}
	We design an elaborate pipeline to automatically retrieve pertinent images from the dataset zoo.
	First, we extract the relevant data cards from the dataset zoo based on the specified task category.
	Next, if the user requests the specific dataset, we compute the fuzzy match scores, quantifying the similarity between the user-designated name and the name of data cards.
	These cards are subsequently reordered in a descending pattern, according to their scores.
	Then the images that meet the requirements are sequentially collected from each dataset.
	Meanwhile, we use WordNet~\cite{miller1995wordnet} to identify semantic relationships between objects.
	
	\subsection{Model Selection}
	\label{sec:model_selection}
	Model selection automates the process of choosing the most suitable model from the model zoo.
	\textbf{Model Zoo.}
	We first construct a comprehensive model zoo where each model is accompanied by a detailed model card and the pre-trained model weights. 
	Each model card incorporates attributes such as the model's name, structure (\eg network depth), parameters, floating point operations per second (FLOPs), inference speed, and performance.
	We have considered AI safety issues when constructing the model zoo. We try to keep the model source open and transparent, and all models are pre-trained with reliable public data.
	\textbf{Model Selection Pipeline.} 
	We also design a elaborate pipeline able to automatically determine the most suitable model in model zoo, tailored to the user's model-specific requirements.
	First, we extract the task-related model cards from the model zoo according to predicated task categorization specified in parsed config.
	If the user has preference to specific model, then calculate the fuzzy matching score between the card's name containing the model structure information and the user's specified requirements.
	Next, we filter out model cards that did not meet the constraints imposed by users regarding the model's parameters, FLOPs, or the inference speed.
	Finally, we select the optimal model from the remaining model cards based on fuzzy matching scores and model performance.
	
	\subsection{Model Training with HPO}
	\label{sec:model_training_hpo}
	Hyperparameter optimization (HPO) or hyperparameter tuning is the problem of choosing a set of optimal hyperparameters for model training, which is crucial for the enhancement of the model performance. 
	Hyperparameters are configuration settings that are not learned during training, but are predefined to control the learning process (\eg learning rate and batch size). AutoMMLab supports various types of HPO baselines, including classic methods (\eg random sampling) and LLM-based methods.
	
	\subsection{Model Deployment}
	\label{sec:model_deployment}
	Model deployment in computer vision means integrating a trained computer vision model into a real-world production environment. 
	Model deployment mainly involves model conversion, graph structure optimization, and model quantization. 
	Our AutoMMLab platform is equipped with MMDeploy~\cite{mmdeploy}, a comprehensive model deployment toolbox, for one-click end-to-end model deployment. It supports a growing list of models that are guaranteed to be exportable to various backends (\eg ONNX, NCNN, OpenVINO). The model converters enable converting OpenMMLab models into backend models that can be run on target devices. 
	The resulting package is also with a inference SDK, supporting data preprocessing, model forward and postprocessing modules during the model inference phase.
	\begin{figure}[t]
		\centering
		\includegraphics[width=0.47\textwidth]{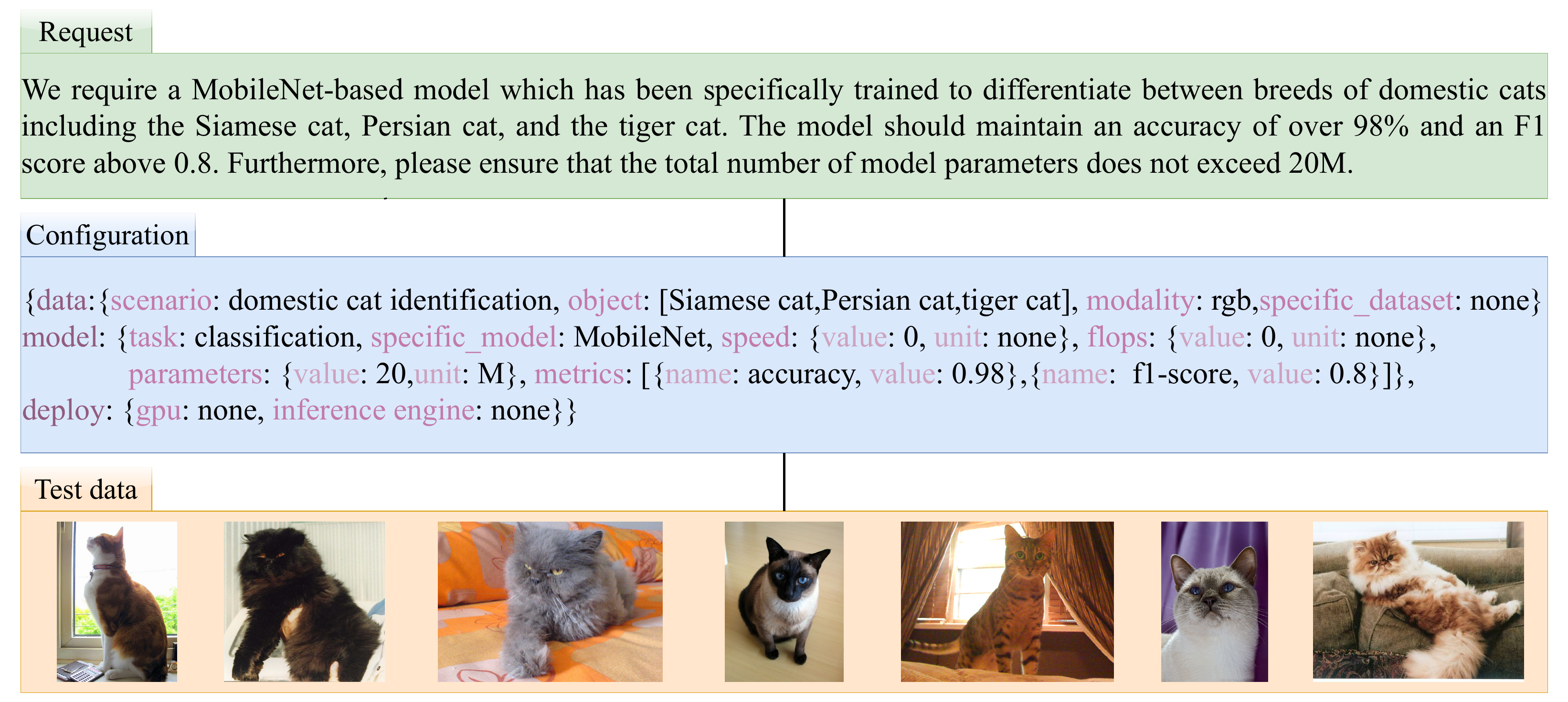}
		\caption{Example of LAMP dataset.} 
		\label{fig:lamp}
	\end{figure}
	
	\section{LAMP Benchmark}
	Large language models (LLMs) have shown amazing emergent abilities in recent studies, however, there lacks a comprehensive benchmark to evaluate the capability of recent LLMs for AutoML. To fill in this blank, we introduce the LAMP (Language-instructed Automated Model Production) benchmark\footnote{This is the v1 version of LAMP, which will be updated as LLMs evolve.}. The benchmark provides a common ground for evaluating and developing the AutoML platform especially for language-instruction based model production. 
	
	\textbf{Testing Data Collection.}
	The benchmark covers four basic computer vision tasks, \ie image classification, object detection, semantic segmentation, and keypoint detection. 
	To ensure the quality of the test data, we ask 5 professional computer vision researchers to manually design 20 diverse CV model requests for each task.  
	In total, we collect 80 unique user instructions and each instruction is accompanied with a ground truth configuration and a test dataset. 
	To enable comparisons of different LLMs for request understanding, we also provide the corresponding ground-truth configuration for each instruction. An example of our test sample is delineated in Figure~\ref{fig:lamp}.
	
	\textbf{Configuration Definition.}
	As shown in Figure~\ref{fig:lamp}, we define three dimensions of model requirements in the  configuration file: ``data'', ``model'', and ``deploy''. 
	``Data'' contains the user's data-related requirements, including: application scenarios, target objects, data modalities, etc.
	``Model'' contains the requirements related to the model, including: task type, model constraints, and the performance that the model is expected to achieve. The constraints of the model may include parameter size, floating point operations per second (FLOPs) and inference speed.
	``Deploy'' contains the user's deployment requirements for the model, including whether to use GPU, which inference engine to use, etc. 
	Note that all values in the configuration can be empty or ``none''. 
	
	We establish three types of evaluation protocols for our LAMP benchmark:
	\begin{itemize}[itemsep=1pt,topsep=0pt,leftmargin=*,itemindent=\parindent]
		\item \textbf{Evaluation of Request Understanding.} \label{sec:evaluation_request_understanding}
		The evaluation of request understanding assesses the ability of LLMs to parse user's requests into configurations, adhering to the predefined JSON format.
		We measure performance based on the accuracy of parsing key-value pairs in the configuration. These pairs can be categorized into item-type pairs and list-type pairs, depending on the data format. In the case of list-type pairs, such as target objects, the pair is considered correctly parsed only if all items in the list are parsed accurately. Note that we apply post-processing and fuzzy matching before calculating the parsing accuracy. 
		Two evaluation metrics are used to gauge the request understanding capabilities: (1) Key-level accuracy calculates the average accuracy of each key-value pair.
		(2) Req-Level accuracy calculates the accuracy of understanding the entire request. 
		\item \textbf{Evaluation of Hyperparameter Optimization.}
		LAMP allows researchers to assess the effectiveness of various Hyperparameter Optimization (HPO) algorithms across a wide range of tasks.
		Different evaluation metrics are utilized for different tasks, with the mean values of the generated model serving as the primary evaluation metric for HPO.
		For image classification, the top-1 accuracy is adopted.
		For object detection, the standard mean Average Precision (mAP) is reported. In the case of semantic segmentation, mean Intersection over Union (mIOU) is employed.
		For keypoint estimation, mAP based on Object Keypoint Similarity (OKS) is used.
		Simultaneously calculate the mean value and standard deviation of four tasks for overall evaluation.
		\item \textbf{End-to-end Evaluation.}
		End-to-end evaluation focuses on the functional quality of the model generated by the request-to-model platform and is assessed using a grading scale ranging from `F' for total failure to `P' for perfect conformity to users' specifications:
		`F' for total failure, scoring 0 points. The platform either fails to generate a functional model.
		`W' for workable model, scoring 1 point. The model is runnable, but it may not fully meet all the users' requirements, such as lower accuracy.
		`P' for perfect model, scoring 2 points. The model perfectly matches the users' expectations.
		There are four types of tasks. The full score for each task is 40, and the total full score is 160. 
	\end{itemize}
	
	\begin{figure*}[t]
		\centering
		\includegraphics[width=0.98\textwidth]{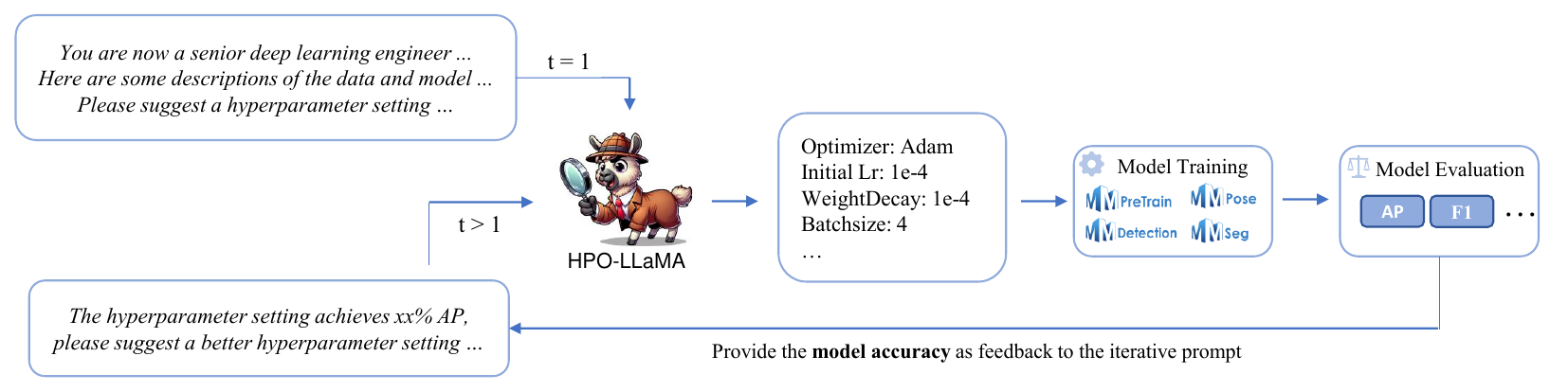}
		\caption{Overview of HPO-LLaMA.
			At the initial step ($t=1$), HPO-LLaMA proposes a hyperparameter configuration based on the description of model and task. Model training is then executed and the training results are passed back to HPO-LLaMA via a text prompt for further rounds ($t>1$).
		} \label{fig:hpo-llama}
	\end{figure*}

	\section{HPO-LLaMA}
	We propose to tackle the challenging HPO task by tuning a LLM-based optimizer, called HPO-LLaMA. 
	Figure~\ref{fig:hpo-llama} demonstrates an overview of our proposed HPO-LLaMA workflow.
	Initially, HPO-LLaMA begins by receiving a nature language description of the HPO problem in conjunction with request-specific details relating to the data and model. Following this, it generates a specific hyperparameter setting tailored to the request.
	Subsequently, HPO-LLaMA acquires the outcomes from the model trainings conducted using the specified hyperparameter setting on the test data and then endeavors to recommend an optimally enhanced hyperparameter setting.
	
	\textbf{Search Space}
	We carefully constructed a compact yet expressive search space for HPO, which includes: optimizer type, initial learning rate, learning rate decay policy, weight decay value, batch size, and the training iterations. 
	Considering the heterogeneity of the four tasks, a bespoke search space is delineated for each individual task.
	
	\textbf{Training HPO-LLaMA}
	We first adopt GPT-assisted data generation method (introduced in Request Understanding) to generate 100 diverse training requests for each of the four CV tasks. 
	And for each training request, we stochastically select 20 distinct hyperparamter settings from the search space.
	Consequently, we extensively execute model training for $8000$ experiments.
	The outcomes of these experiments are compiled into a substantial dataset, comprising triplets of "request-hyperparameter-performance".
	we further conduct a comprehensive evaluation to elucidate the correlation between hyperparameters and model performance.
	This is primarily to substantiate the appropriateness of the predetermined selection for hyperparameter space.
	Please refer to supplementary for details.
	
	The triplets of ``request-hyperparameter-performance'' contains rich knowledge about model training and hyperparameter tuning. 
	We utilize the triplet data to construct both single-round and multi-round dialogues, thereby empowering HPO-LLaMA to activate the capability of predicting efficient hyperparameters from these dialogues.
	Specifically, when constructing the first round of dialogue, the contextual prompts includes the hyperparameter search space, along with a detailed description of the data and the model. The hyperparameter settings to be predicted are derived from the top-$k$ optimal hyperparameter settings corresponding to the request.
	In multiple rounds of conversations, the contextual prompt provided to HPO-LLaMA includes historical prompt of previous conversations and the performance of the model trained on the test set for the last time the model predicted hyperparameters.
	The predicted hyperparameter settings are derived from top-$k$ of all hyperparameter settings corresponding to the request and are the best performing hyperparameter setting in current multi-round dialogue.
	Finally, we construct the training dataset for up to three dialogue rounds and selectively sample a subset from the entire pool of potential multi-turn conversation data.
	We then use LoRA to fine-tune LLaMA-7B and get our HPO-LLaMA.
	Considering the limited volume of data in a single-round dialogue, we mix dialogue rounds ranging from one to three together to train the HPO-LLaMA.

	\section{Experiments}
	\subsection{Request Understanding}
	\textbf{Baseline Models.}
	We built several baseline methods by giving LLMs carefully designed prompts.
	The prompts comprise an instruction and, optionally, a few demonstrations of the anticipated behavior.
	LLaMA2-7B without instruction fine-tuning cannot accurately follows instructions, so we choose LLaMA2-7B-Chat with instruction fine-tuning as our baseline.
	As GPTs are currently the most successful LLMs for general language tasks, we opt to employ GPT-3.5-turbo and GPT-4 through API calls as baseline models to evaluate the request understanding ability. We also evaluate the request understanding ability of PaLM2.
	
	\textbf{Main Results.}
	Table~\ref{tab:RU} presents the outcomes of various methods evaluated on LAMP 
	utilizing the evaluation metrics delineated in the section of LAMP Benchmark.
	The findings manifest that LLaMA2-7B-Chat, which did not use the dataset we constructed for fine-tuning, is not able to completely correctly parse the request.
	Both RU-LLaMA and GPT4 demonstrate exciting capability in understanding requests.
	RU-LLaMA outperforms PaLM2, GPT-3.5-turbo, and GPT-4 with a 7B-parameter model for both key-level and req-level evaluations.
	\begin{figure*}[htbp]
		\centering
		\begin{subfigure}[b]{0.246\textwidth}
			\includegraphics[width=\textwidth]{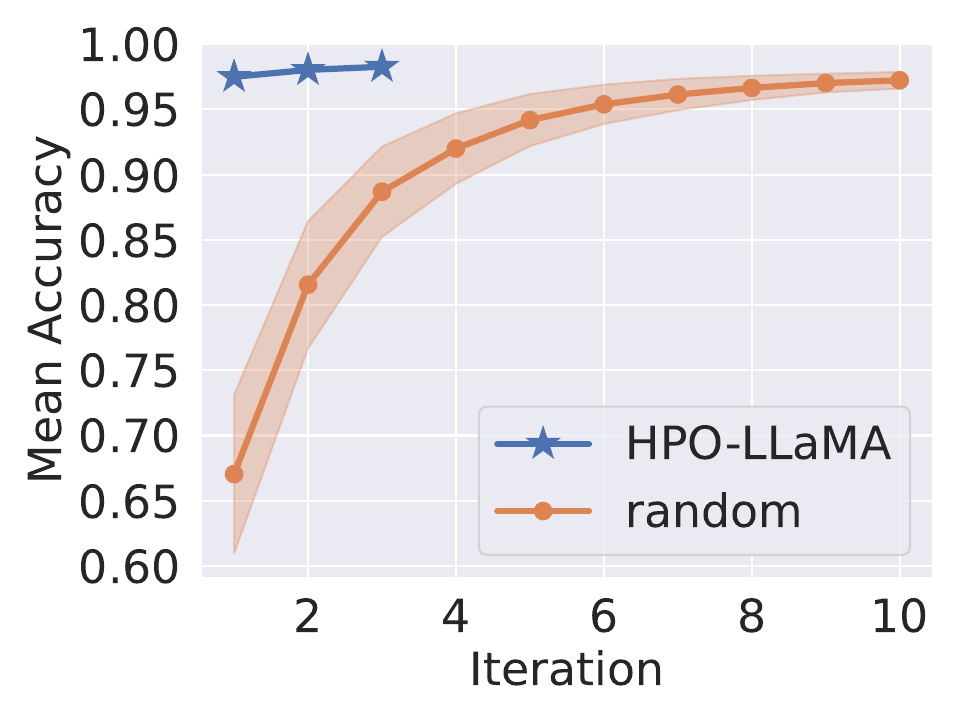}
			\caption{Image Classification}
		\end{subfigure}
		\begin{subfigure}[b]{0.246\textwidth}
			\includegraphics[width=\textwidth]{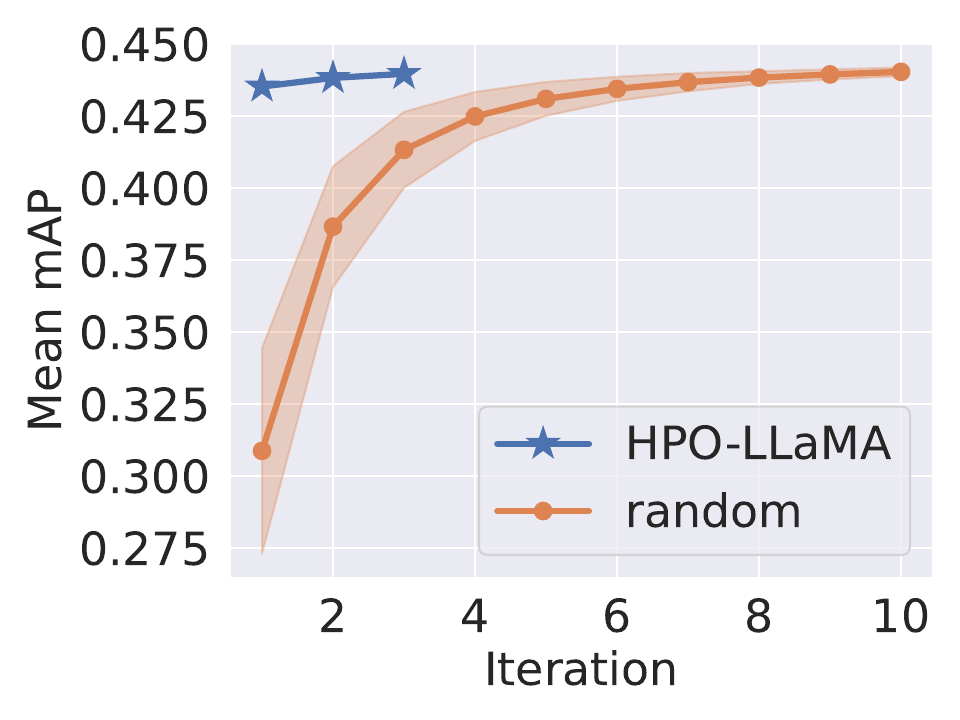}
			\caption{Object Detection}
		\end{subfigure}
		\begin{subfigure}[b]{0.246\textwidth}
			\includegraphics[width=\textwidth]{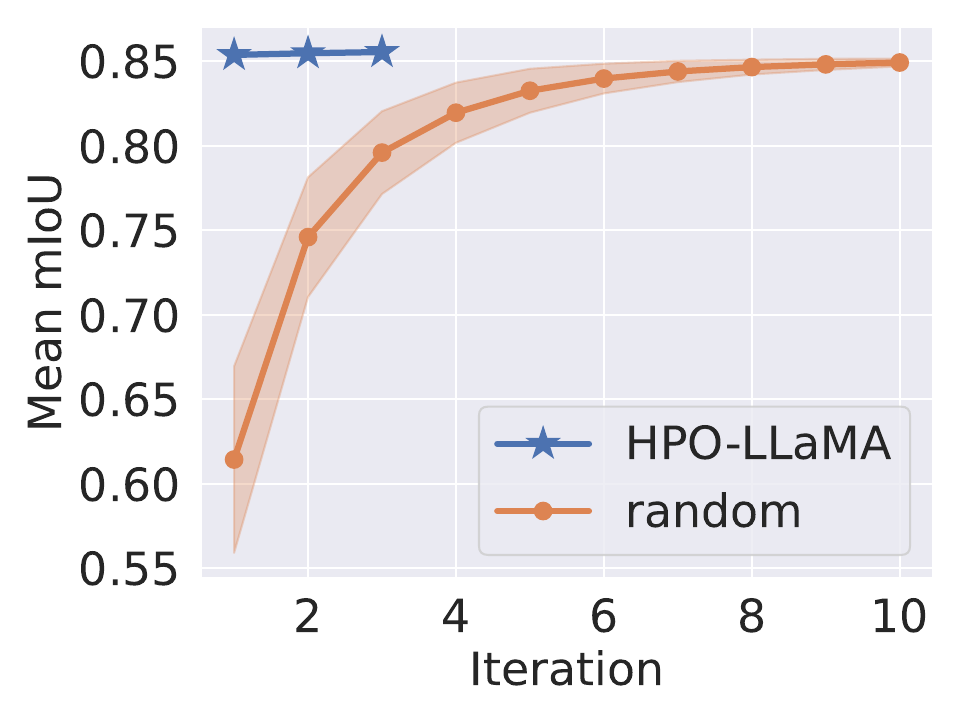}
			\caption{Semantic Segmentation}
		\end{subfigure}
		\begin{subfigure}[b]{0.246\textwidth}
			\includegraphics[width=\textwidth]{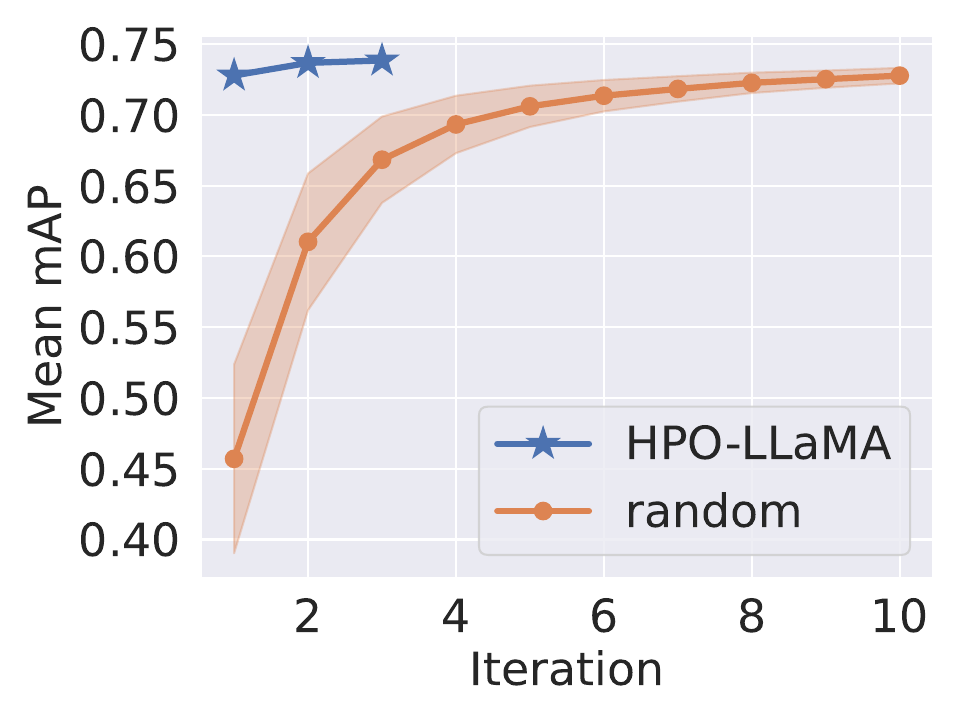}
			\caption{Keypoint Detection}
		\end{subfigure}
		\caption{HPO results of HPO-LLaMA and random sampling baselines on four tasks
			: (a) image classification, (b) object detection, (c) semantic segmentation and (d) keypoint detection.
			HPO-LLaMA demonstrates significantly higher efficiency.
		} 
		\label{fig:main}
		\label{fig:exp_hpo-llama}
	\end{figure*}
	\begin{table}[ht]
		\centering
		\setlength{\tabcolsep}{1mm}
		\begin{tabular}{ccccc}
			\toprule
			\multirow{2}{*}{Model} & \multicolumn{3}{c}{Key-Level} & \multirow{2}{*}{\begin{tabular}[c]{@{}c@{}}Req-Level\\ \end{tabular}} \\ \cline{2-4}
			& Item & List & Total & \\ 
			\midrule
			LLaMA2-7B-Chat  & 85.71 & 50.00 & 77.78 & 0 \\
			PaLM2           & 96.79 & 88.13 & 94.86 & 63.75 \\
			GPT-3.5-turbo   & 96.43 & 95.63 & 96.25 & 72.50 \\
			GPT-4      & 97.50 & 93.13  & 96.53 & 80.00  \\
			\rowcolor{ourscolor}
			RU-LLaMA     & \textbf{98.57} & \textbf{96.88} & \textbf{98.20}  & \textbf{86.25} \\
			\bottomrule
		\end{tabular}
		\caption{Evaluation of request understanding (RU). Best results are marked in bold.}
		\label{tab:RU}
	\end{table}
	
	\begin{table*}[htbp]
		\centering
		\begin{tabular}{cc|ccccc}
			\toprule
			Model & \#R & Cls. & Det. & Seg. & Kpt.  \\ 
			\midrule
			BayesianRF  & 5  & 0.618$\pm$0.287 & 0.291$\pm$0.211 & 0.847$\pm$0.044 & 0.069$\pm$0.136 &  \\
			BayesianGP  & 5  & 0.761$\pm$0.264 & 0.280$\pm$0.208 & 0.848$\pm$0.041 & 0.081$\pm$0.150 &  \\
			LLaMA2-7B  & 1  & 0.839$\pm$0.213 & 0.128$\pm$0.164 & 0.291$\pm$0.409 & 0$\pm$0 &  \\
			PaLM2           & 1  & 0.964$\pm$0.056 & 0.367$\pm$0.196 & 0.845$\pm$0.067 & 0.719$\pm$0.079     \\
			GPT-3.5-turbo   & 1  & 0.849$\pm$0.214   & 0.364$\pm$0.194  & 0.852$\pm$0.044  & 0.204$\pm$0.160    \\
			GPT-4           & 1  & 0.861$\pm$0.188   & 0.434$\pm$0.147  & 0.803$\pm$0.194  & 0.096$\pm$0.158   \\
			\rowcolor{ourscolor}
			HPO-LLaMA       & 1  & \underline{0.975$\pm$0.028}    & \underline{0.435$\pm$0.148}    & \underline{0.854$\pm$0.042}    & \underline{0.728$\pm$0.051}   \\ 
			\rowcolor{ourscolor}
			HPO-LLaMA       & 3  & \textbf{0.983$\pm$0.020}     & \textbf{0.440$\pm$0.150}       & \textbf{0.856$\pm$0.043}       & \textbf{0.738$\pm$0.053}   \\ \bottomrule
		\end{tabular}
		\caption{Evaluation of HPO. \#R means the number of iteration rounds. The mean and standard deviation on for task are exhibited. Best results are marked in bold, second best results are underlined.}
		\label{tab:hpo}
	\end{table*}
	
	\subsection{Hyper-parameter Optimization (HPO)}
	\textbf{Baselines:} \textbf{(1) Random Sampling.} In the realm of HPO, random sampling is usually employed as an important baseline. Specifically, we uniformly sample values from our pre-defined hyperparameter search space. We perform sampling for 10 rounds and subsequently identify the optimal configuration to serve as our baselines. 
	Nevertheless, we observed considerable variance across individual trials resulting from this sampling approach. 
	To address this, we repeated the 10-round process 1,000 times and reported the average and standard deviation.
	\textbf{(2) Bayesian.} We compare with two traditional Bayesian HPO methods, using random forest (BayesianRF) and Gaussian process (BayesianGP) as surrogate models. For each method we perform five rounds of optimization.
	\textbf{(3) LLM prompting.}
	We compare with several popular LLMs including LLaMA2-7B-Chat, PaLM2, GPT-3.5-turbo, and GPT-4. Specifically, we directly prompt these general-purpose LLMs with some few-shot examples to produce hyperparameter settings. Note that we used the same prompt as our HPO-LLaMA for fair comparisons. 
	
	\textbf{Main Results.}
	Figure~\ref{fig:exp_hpo-llama} shows that our proposed HPO-LLaMA demonstrates a remarkable ability for hyperparameter optimization. It significantly outperforms the random sampling baseline on all four representative tasks. It is worth noting that with only one single round trial, our HPO-LLaMA already produces good hyper-parameters. 
	In Table~\ref{tab:hpo}, we compare our proposed HPO-LLaMA with Bayesian HPO methods using a 5-round evaluation protocol, as well as several popular LLMs with a 1-round evaluation protocol, demonstrating the effectiveness of our approach.
	In our experiment, we observed that relying on the few-shot capability of LLaMA2-7B-Chat to suggest hyperparameters doesn't consistently yield the anticipated output, and the effectiveness of the suggested hyperparameters is poor.
	An intriguing finding is that after meticulously designing prompts for GPT-3.5-turbo and GPT-4, they exhibit commendable abilities in proposing hyperparameter settings in tasks relating to image classification, object detection, and semantic segmentation. However, in contrast, for the relatively uncommon task of keypoint detection, they fail to suggest useful hyperparameter settings.
	The HPO-LLaMA model proposed in our study demonstrates the capacity to provide useful hyperparameter recommendations across all four computer vision tasks, and achieves the best performance.
	At the same time, as the number of rounds increases, HPO-LLaMA can continue to search for better hyperparameter settings.
	\begin{table}[htbp]
		\centering
		\setlength{\tabcolsep}{0.8mm}
		\begin{tabular}{cc|ccccc}
			\toprule
			RU              & HPO & Cls. & Det. & Seg. & Kpt. & Total \\ 
			\midrule
			LLaMA2-7B  & LLaMA2-7B & 0     &0     & 0   & 0   & 0 \\
			PaLM2           & PaLM2          & 14    &25    & 27  & 15  &81    \\
			GPT-3.5-turbo   & GPT-3.5-turbo        & 24    &24    &25   &11   &84    \\
			GPT-4           & GPT-4          & 17    &27    &29   &14   &87    \\
			\rowcolor{ourscolor}
			RU-LLaMA        & HPO-LLaMA  &\textbf{31}    &\textbf{31}     &\textbf{32}   &\textbf{18}   &\textbf{112}  \\ \bottomrule
		\end{tabular}
		\caption{End-to-end evaluation on LAMP benchmark. The assessment is based on a grading system scoring from `0' to `2', where `0' denotes `total failure', `1' denotes `workable model', 2 denotes `perfect model'. The full score for each task is 40, and the total full score is 160.}
		\label{tab:end2end}
	\end{table}
	
	\subsection{End-to-end Evaluation}
	Table~\ref{tab:end2end} reports the results of end-to-end evaluation on LAMP benchmark. There are four types of model training tasks, \ie image classification (Cls.), object detection (Det.), semantic segmentation (Seg.) and keypoint detection (Kpt.). 
	We employ various LLMs for request understanding (RU) and hyperparameter optimization (HPO) respectively.
	We find that our AutoMMLab empowered by RU-LLaMA and HPO-LLaMA significantly outperform GPT-3.5 and GPT-4 with a 7B-parameter model, scoring 112/160 in the overall evaluation and achieves the best performance on all four computer vision tasks.
	
	\section{Conclusion}
	This paper presents AutoMMLab, an innovative AutoML system that harnesses the power of LLMs to fully automate the model development process for computer vision tasks, guided by natural language instructions.
	We develop RU-LLaMA for understanding the user's requests and scheduling the whole workflow.
	Furthermore, we propose HPO-LLaMA, a novel LLM-empowered method for effective hyperparameter optimization.
	Additionally, we present the LAMP (Language-instructed Automated Model Production) Benchmark to facilitate the evaluation of end-to-end prompt-based model production.
	We equip our AutoMMLab with various LLMs and test on LAMP, demonstrating the superiority of our proposed RU-LLaMA and HPO-LLaMA.
	We hope our work could have broad applicability and inspire further research in related areas.
	
	\bibliography{main}
	
	\clearpage
	\appendix
	\setcounter{table}{0}
	\renewcommand{\thetable}{A\arabic{table}}
	\setcounter{figure}{0}
	\renewcommand{\thefigure}{A\arabic{figure}}
	\setcounter{section}{0}
	\renewcommand{\thesection}{A\arabic{section}}
	
	\section{Request Understanding}
	\subsection{Request Generation Prompts}
	We carefully design prompts for GPT4 and subsequently refine the prompts iteratively to generate requests in batches. Table~\ref{tab:ru_promt_request_base} illustrates the prompt template we designed, which mainly consists of basic prompts, constraints on generated content and anticipated examples. Table~\ref{tab:ru_prompt_request_constraints} shows the constraints we designed.
	Table~\ref{tab:ru_prompt_request_example} exhibits three examples of generated requests.
	\begin{table}[htbp]
		\centering
		\caption{Request generation prompts template.}
		\resizebox{0.49\textwidth}{!}{
			\begin{tabular}{p{0.49\textwidth}}
				\hline
				You are now a client who specifically provides requirements to the project manager from a computer vision model development team, and you are asked to provide 10 diverse requirements, each requirement is prefixed with \#\#\#requirement\#\#\#. You must selectively describe what function and performance the model you want can achieve in what application scenarios. \\
				The following are the conditions that need to be met for the requirements you provide$:$\\
				\{\{Request Generation Constraints\}\} \\
				Here are some examples you can refer to:\\
				\{\{Request Generation Examples\}\}\\
				\hline
			\end{tabular}
		}
		\label{tab:ru_promt_request_base}
	\end{table}
	\begin{table}[hbtp]
		\centering
		\caption{Examples of request.}
		\resizebox{0.49\textwidth}{!}{
			\begin{tabular}{p{0.49\textwidth}}
				\hline
				\#\#\#requirement\#\#\#Could you assist in finding a model capable of detecting the license plate in the car pictures captured by the camera at the highway intersection, and the model should be deployable on a GPU and the maximum FLOPs of the model should not exceed 20 GFLOPs. The model's detection accuracy is required to be more than 95\%. \\ 
				\#\#\#requirement\#\#\#Please help me implement an image segmentation model. The mIOU of the model on the pascal voc 2012 dataset is not less than 0.8, and the number of model parameters does not exceed 5 billion. The model should be deployed using ncnn for efficient inference and be lightweight enough to run on a standard laptop without requiring a GPU. \\
				\#\#\#requirement\#\#\#I often see all kinds of beautiful flowers in my daily life, but I can't recognize their names. Can you help me realize a model that can recognize common flowers? The accuracy of the model should be above 90\%. The speed at which the model predicts a picture cannot exceed 3 seconds. \\
				\hline
			\end{tabular}
		}
		\label{tab:ru_prompt_request_example}
	\end{table}
	\begin{table}[tbp]
		\centering
		\caption{Request generation constraints.}
		\resizebox{0.46\textwidth}{!}{
			\begin{tabular}{p{0.49\textwidth}}
				\hline
				1. Try not to repeat verbs in each requirement, and maximize the diversity of requirement expressions.\\
				2. The tone of expression for requirements should be as diverse as possible.\\
				3. The requirements are written in English, the requirement should to be a paragraph, imperative sentences are allowed.\\
				4. The required application scenarios should be diversified, including various application scenarios of computer vision, such as: smart agriculture, smart factories, smart life, smart education, face recognition, automatic driving, medical imaging, etc.\\
				5. The modality of the image may be different under different requirement scenarios, for example: RGB image, grayscale image, infrared image, depth image, radar image, CT image and MRI image, etc. The image modality should match the corresponding requirement scenario.\\
				6. Specific public datasets, such as COCO, ImageNet, etc., can be specified in the requirements, or not specified.\\
				7. The tasks should be diverse, including various types of computer vision tasks. The types of tasks you generate should be in the following types: image classification, object detection, image segmentation, keypoint detection.\\
				8. Common computer vision models can be specified in the requirements, such as: ResNet50, FasterRCNN, YOLO, DETR, etc., or not specified.\\
				9. In the requirements, information such as the maximum parameter amount of the model, the maximum FLOPs of the model, and the speed required for the model to infer a sample can be constrained.\\
				10. The requirements need to specify the way to evaluate the performance of the model, such as: accuracy, F1, AUC, mAP and other commonly used model evaluation indicators in computer vision tasks, and the value that the model is expected to achieve.\\
				11. The deployment environment of the model can be specified in the requirements, for example: whether GPU is required or not.\\
				12. The inference engine of the deployment environment can be specified, for example: onnx runtime, OpenVINO, ncnn.\\
				\hline
			\end{tabular}
		}
		\label{tab:ru_prompt_request_constraints}
	\end{table}
	
	\begin{table}[htbp]
		\centering
		\caption{Example of request parsing.}
		\resizebox{0.49\textwidth}{!}{
			\begin{tabular}{p{0.49\textwidth}}
				\hline
				\#\#\#requirement\#\#\#I am interested in developing a smart agriculture system that can classify different types of crops in the field using drone-captured RGB images. The model should be able to classify common crops with an Accuracy of 0.75 or higher, and the model should be able to infer a sample within 500 GFLOPs. The model should be deployed using ncnn for efficient inference and be lightweight enough to run on a standard laptop without requiring a GPU. \\
				\#\#\#parse\#\#\#
				\{``data'': \{``description'': ``Drone-captured RGB images of crops in the field, the dataset contains common crops.'', ``scenario'': ``agriculture'', ``object'': ["crops"], ``modality'': ``rgb'', ``specific'': []\}, ``model'': \{``description'': ``A model that can classify common crops with an Accuracy of 75\% or higher.'', ``task'': ``classification'', ``specific\_model'': ``none'', ``speed'': \{``value'': 0, ``unit'': ``none''\}, ``flops'': \{``value'': 500, ``unit'': ``GFLOPs''\}, ``parameters'': \{``value'': 0, ``unit'': ``none''\}, ``metrics'': [\{``name'': ``accuracy'', ``value'': 0.75\}]\}, ``deploy'': \{``description'': ``Standard laptop without requiring a GPU.'', ``device'': ``cpu'', ``inference engine'': ``ncnn''\}\} \\
				\hline
		\end{tabular}}
		\label{tab:ru_prompt_parse_example}
	\end{table}
	
	\subsection{Request Parsing Prompts}
	Table~\ref{tab:ru_prompt_parse_base} shows the prompt template we designed for request parsing, which consists of three parts: basic prompt, JSON format definition of the parsed configuration, and expected examples.
	Listing~\ref{list:config_json_format} shows the JSON format of the configuration we defined.
	Table~\ref{tab:ru_prompt_parse_example} presents an example of request parsing.
	This prompts template is used in multiple places, including: using GPT-4 to generate training data for request understanding, building the baseline for request understanding, and serving as context information when fine-tuning RU-LLaMA.
	
	\begin{table*}[htbp]
		\centering
		\caption{Request parsing prompts template.}
		\resizebox{0.96\textwidth}{!}{
			\begin{tabular}{p{0.96\textwidth}}
				\hline
				You are a product manager of a professional computer vision model development team. The user will use a \#\#\#requirement\#\#\# to describe the model he wants your team to help him train and deploy. You need to summarize the user's requirements for data, models and deployment based on the user's description. \\
				For the user's data requirements, you need to summarize: the user's application scenario, the target object in the image, the data modality, and the dataset required by the user, etc. 
				For the user's model requirements, you need to summarize: the tasks that the user wants to achieve, the model that the user specifies to use, the running speed of the model, the number of parameters of the model, the amount of calculation of the model, the measurement method and target value of the model performance. 
				For the user's deployment requirements, you need to summarize: information such as the deployment environment and equipment required by the user. 
				You need to pay attention that for tasks that require both detection and classification, they need to be divided into detection tasks. \\
				For the user's deployment requirements, you need to summarize: information such as the deployment environment and equipment required by the user. \\
				Finally, you need to return \#\#\#parse\#\#\# that conforms to the following json specification: \\
				\{\{JSON Format of Configuration\}\} \\
				Here are some cases you can refer to: \\
				\{\{Request Parsing Examples\}\} \\
				\hline
		\end{tabular}}
		\label{tab:ru_prompt_parse_base}
	\end{table*}
	
	\newpage
	\begin{lstlisting}[basicstyle=\tiny,caption={JSON format of configuration.},label={list:config_json_format}]
		{
			"data": {
				"type": "object",
				"description": "User's requirements for dataset.",
				"properties": {
					"description": {
						"type": "string",
						"description": "Detailed description of user data requirements."
					},
					"scenario": {
						"type": "string",
						"description": "Application scenario of user."
					},
					"object": {
						"type": "array",
						"items": {
							"type": "string"
						},
						"description": "Target objects that user wants to identify, classify, detect or segment."
					},
					"modality": {
						"type": "string",
						"description": "Datasets modality for user application scenario."
					},
					"specific": {
						"type": "array",
						"items": {
							"type": "string"
						},
						"description": "The specific dataset specified by the user."
					}
				}
			},
			"model": {
				"type": "object",
				"description": "User's requirements for model.",
				"properties": {
					"description": {
						"type": "string",
						"description": "Detailed description of user model requirements."
					},
					"task": {
						"type": "string",
						"enum": [
						"classification",
						"detection",
						"segmentation",
						"keypoint"
						],
						"description": "The task that the user wants model to accomplish."
					},
					"specific_model": {
						"type": "string",
						"description": "The specific model that the user wants to implement the target task."
					},
					"speed": {
						"type": "object",
						"properties": {
							"value": {
								"type": "number",
								"description": "Value of speed . Default: 0"
							},
							"unit": {
								"type": "string",
								"enum": [
								"ms",
								"s",
								"min",
								"h",
								"fps",
								"none"
								],
								"description": "Unit of speed. Default: none"
							}
						},
						"description": "The speed at which the user requires the model to infer a data, unit in seconds. Default: 0"
					},
					"flops": {
						"type": "object",
						"properties": {
							"value": {
								"type": "number",
								"description": "Value of floating point operations number. Default: 0"
							},
							"unit": {
								"type": "string",
								"enum": [
								"FLOPs",
								"MFLOPs",
								"GFLOPs",
								"TFLOPs",
								"PFLOPs",
								"EFLOPs",
								"none"
								],
								"description": "Unit of floating point operations number. Default: none"
							}
						},
						"description": "Floating point operations number of model."
					},
					"parameters": {
						"type": "object",
						"properties": {
							"value": {
								"type": "number",
								"description": "Value of parameter number. Default: 0"
							},
							"unit": {
								"type": "string",
								"enum": [
								"K",
								"M",
								"B",
								"none"
								],
								"description": "Unit of arameter number. Default: none"
							}
						},
						"description": "Parameter number of model."
					},
					"metrics": {
						"type": "array",
						"items": {
							"type": "object",
							"properties": {
								"name": {
									"type": "string",
									"description": "metirc type."
								},
								"value": {
									"type": "number",
									"description": "metirc value."
								}
							},
							"description": "User's requirements for model performance. There may be multiple emetircs, and each metirc corresponds to a metirc indicator and the desired value."
						}
					}
				}
			},
			"deploy": {
				"type": "object",
				"description": "User's requirements for deploy environment.",
				"properties": {
					"description": {
						"type": "string",
						"description": "Detailed description of user deploy environment requirements."
					},
					"device": {
						"type": "string",
						"enum": [
						"cpu",
						"gpu",
						"none"
						],
						"description": "The deployment environment is GPU-accelerated or CPU-only or not specified. Default: none."
					},
					"inference engine":{
						"type": "string",
						"enum": [
						"onnxruntime",
						"ncnn",
						"openvino",
						"none"
						],
						"description": "Deployment environment's inference engine. Default: none."
					}
				}
			}
		}
	\end{lstlisting}

	\section{Dataset Zoo}
	We employ a variety of publicly available datasets to constitute the dataset zoo, designed to accommodate numerous tasks and diverse scenarios.
	For example, ImageNet1k~\cite{krizhevsky2017imagenet} hosts more than 10 million images spanning across 1,000 categories like animals, plants and vehicles; COCO~\cite{lin2014microsoft} embraces over 200,000 images and encapsulating location annotations of 80 frequently encountered categories in daily life; Cityscapes~\cite{cordts2016cityscapes} holds more than 5,000 images supplied with pixel-level annotations for pedestrians, vehicles, roads, and buildings in urban street scenes; AP10K~\cite{yu2021ap10k} has 10,000 images with a variety of animal key point annotations.
	Our dataset zoo supports the continuous expansion of the datasets list, facilitating its applicability to a wider range of tasks and application scenarios.
	
	\section{Model Zoo}
	Table~\ref{tab:model_zoo} delineates the models currently included by our model zoo.
	Our model zoo also supports a growing list of pretrained models for different tasks.
	Regarding image classification, our model zoo contains popular CNN models (\eg ResNet~\cite{he2016deep} and EfficientNet~\cite{tan2019efficientnet}), Vision Transformers (\eg ViT~\cite{dosovitskiy2020image}), etc. 
	In terms of object detection, the object detection frameworks contained in our model zoo include: two-stage RCNN series~\cite{girshick2015fast, ren2015faster}, one-stage YOLO series~\cite{redmon2018yolov3, yolox2021}, DETR-based detectors~\cite{zhu2021deformable}, etc. 
	Concerning semantic segmentation, we have DeepLab~\cite{chen2017rethinking, deeplabv3plus2018}, UPerNet~\cite{xiao2018unified}, Segformer~\cite{xie2021segformer}, etc.
	For pose estimation, we have SimpleBaseline2D~\cite{xiao2018simple}, HRNet~\cite{sun2019deep}, LiteHRNet~\cite{yu2021lite}, etc.
	
	\begin{table}[htbp]
		\centering
		\caption{Models included in model zoo.}
		\resizebox{0.49\textwidth}{!}{
			\begin{tabular}{|>{\centering\arraybackslash}p{0.1\textwidth}|>{\centering\arraybackslash}p{0.39\textwidth}|}
				\hline
				Task & Models \\ \hline
				Classification & VGG~\cite{simonyan2014very}, Inception~\cite{szegedy2016rethinking}, ResNet~\cite{he2016deep}, SEResNet~\cite{hu2018squeeze}, EfficientNet~\cite{feurer2015efficient}, MobileNet~\cite{Howard_2019_ICCV}, MobileViT~\cite{mehta2021mobilevit}, ShuffleNet~\cite{zhang2018shufflenet,ma2018shufflenet}, ViT~\cite{dosovitskiy2021an}, SwinTransformer~\cite{liu2021Swin}  \\ \hline
				Objection Detection & FSAF~\cite{zhu2019feature}, FCOS~\cite{tian2019fcos}, SSD~\cite{Liu_2016}, RetinaNet~\cite{lin2017focal}, Faster-RCNN~\cite{Ren_2017}, YOLOv3~\cite{redmon2018yolov3}, YOLOX~\cite{yolox2021}, DETR~\cite{detr} \\ \hline
				Senmantic Segmentation & FCN~\cite{shelhamer2017fully}, PSPNet~\cite{zhao2017pspnet}, DeepLab~\cite{chen2017rethinking, deeplabv3plus2018}, UNet~\cite{ronneberger2015u}, APCNet~\cite{He_2019_CVPR}, BiSeNet~\cite{yu2018bisenet, yu2021bisenet}, FastFCN~\cite{wu2019fastfcn}, UperNet~\cite{xiao2018unified}, DANet~\cite{fu2018dual}, Segformer~\cite{xie2021segformer} \\ \hline
				Keypoint Detection & SimpleBaseline2D~\cite{xiao2018simple}, HRNet~\cite{sun2019deep}, LiteHRNet~\cite{yu2021lite} \\
				\hline
		\end{tabular}}\label{tab:model_zoo}
	\end{table}
	
	\section{Hyperparameter Optimization}
	\subsection{Search Space}
\begin{figure*}[htbp]
		\centering
		\includegraphics[width=0.98\textwidth]{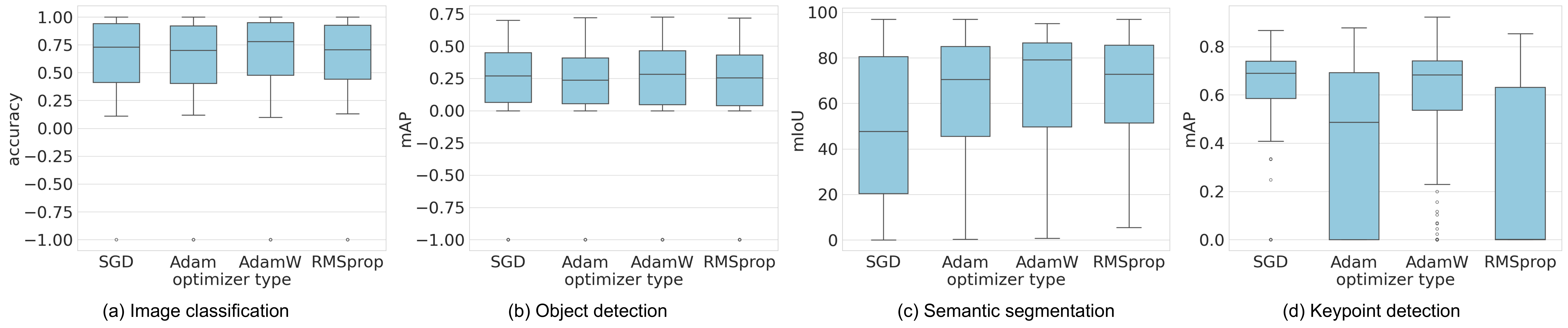}
		\caption{Boxplots for different optimizer types.} \label{fig:cor_opt}
		\vspace{1cm}
		\includegraphics[width=0.98\textwidth]{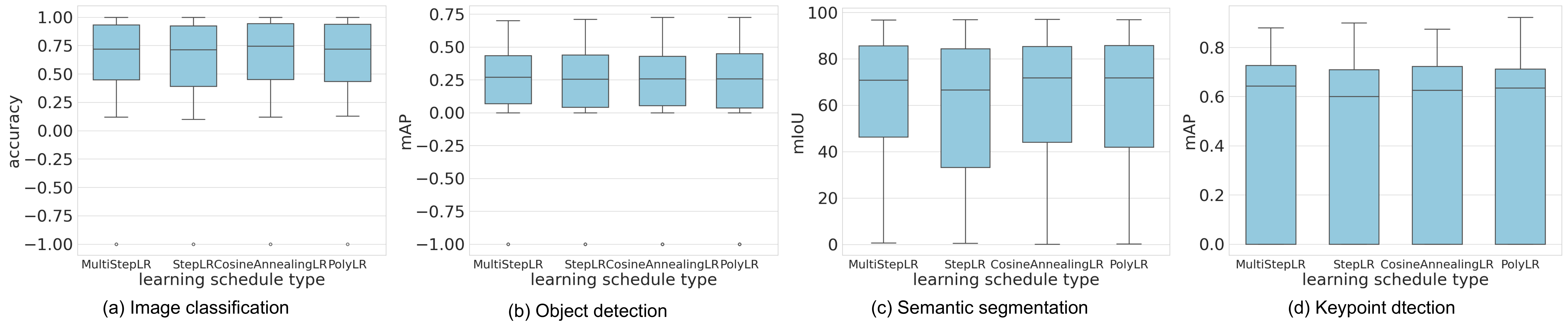}
		\caption{Boxplots for different learning schedule types.} \label{fig:cor_lrs}
	\end{figure*}

	Table~\ref{tab:search_space} details the hyperparameter search space we set for four computer vision tasks.
	We further analyze the correlation between the hyperparameters and model performance in the 8000 trained models. 
	For numerical hyperparameters, we use Pearson correlation coefficients for quantitative analysis.
	And for categorical hyperparameters, we use boxplots for qualitative analysis.
	The results are shown in the Table~\ref{tab:search_space_corr}, Figure~\ref{fig:cor_opt}, and Figure~\ref{fig:cor_lrs}.
	It can be seen from the charts that the performance of the model has larger correlation with the initial learning rate and optimizer type. Models trained with the AdamW optimizer are more likely to perform better on four computer vision tasks.
	\begin{table*}[htbp]
		\centering
		\caption{Hyperparameter search space.}
		\resizebox{0.98\textwidth}{!}{
			\begin{tabular}{|>{\centering\arraybackslash}p{0.2\textwidth}|>{\centering\arraybackslash}p{0.20\textwidth}|>{\centering\arraybackslash}p{0.20\textwidth}|>{\centering\arraybackslash}p{0.20\textwidth}|>{\centering\arraybackslash}p{0.20\textwidth}|}
				\hline
				& Classification             & Object detection             & Semantic segmentation             & Keypoint detection             \\ \hline
				Optimizer types                & \multicolumn{4}{c|}{SGD~\cite{sutskever2013importance}, Adam~\cite{kingma2017adam}, AdamW~\cite{loshchilov2019decoupled}, RMSprop~\cite{graves2014generating}}                            \\ \hline
				Learning rate decay strategies & \multicolumn{4}{c|}{MultiStepLR, ConsineAnnealingLR, StepLR, PolyLR}      \\ \hline
				Initial learning rate range    & \multicolumn{4}{c|}{$[10^{-8},0.1]$}                  \\ \hline
				Weight decay range             & \multicolumn{4}{c|}{$[10^{-5},0.1]$}                  \\ \hline
				Training iteration range       & {[}2000,5000{]} & {[}4000,9000{]} & {[}2000,7000{]} & {[}2000,5000{]} \\ \hline
				Batch size range               & {[}1,64{]}      & {[}1,16{]}      & {[}2,8{]}       & {[}2,64{]}      \\ \hline
		\end{tabular}}
		\vspace{1cm}
		\caption{Correlation between numerical hyperparameters and the final performance of the model. The assessment is based on Pearson correlation coefficient scoring from `-1' to `1', where `-1' denotes `strong negative correlation', `1' denotes `strong positive correlation', and `0' denotes `no correlation'.}
		\begin{tabular}{|>{\centering\arraybackslash}p{0.2\textwidth}|>{\centering\arraybackslash}p{0.13\textwidth}|>{\centering\arraybackslash}p{0.16\textwidth}|>{\centering\arraybackslash}p{0.19\textwidth}|>{\centering\arraybackslash}p{0.16\textwidth}|}
			\hline
			& Classification & Object detection & Semantic segmentation & Keypoint detection \\ \hline
			Initial learning rate & -0.22          & -0.49            & 0.17                  & -0.43              \\ \hline
			Weight decay          & -0.11          & 0.01             & -0.14                 & -0.07              \\ \hline
			Training iteration    & 0.09           & -0.01            & 0.10                  & -0.01              \\ \hline
			Batch size            & 0.08           & 0.12             & 0.01                  & -0.01              \\ \hline
		\end{tabular}
		\label{tab:search_space_corr}
		\label{tab:search_space}
	\end{table*}
	
	\subsection{HPO Prompts}
	Table~\ref{tab:hpo_prompt_template} presents the template of our hyperparameter optimization prompts. 
	Listing~\ref{list:hpo_json_format} shows the hyperparameter search space described in JSON format.
	Listing~\ref{list:hpo_example_datamodel} shows the information of data and model described in JSON format.
	Listing~\ref{list:hpo_example_hp} shows the hyperparameter setting described in JSON format.
	The HPO prompts serve as the hints in the development of the baseline for the hyperparameter optimization model employing Generative Pre-trained Transformers (GPTs).
	Additionally, they function as contextual inputs during the training process of HPO-LLaMA.
	\begin{table*}[h]
		\centering
		\caption{Hyperparameter optimization prompts template.}
		\begin{tabular}{p{0.96\textwidth}}
			\hline
			\#\#\# Round 1 \#\#\# \\
			You are now a senior deep learning engineer assistant. User will give you some json descriptions of the deep learning model and training data.
			Please provide the best set of hyperparameters for training this model to the user. The given hyperparameters need to conform to the following json format: \\
			\{\{Hyperparameter search space definition in JSON format\}\} \\
			In multiple rounds of conversations, the user will train the model based on the hyperparameters provided by the assistant and tell the assistant the results of the trained model on the test dataset.
			The assistant needs to think and reason to provide a better set of hyperparameters so that the model trained using these hyperparameters can achieve better results on the test dataset.\\
			\\
			User:\{\{Data and model description in JSON format\}\} \\
			Assistant:\{\{Hyperparameter setting in JSON format\}\} \\
			\\
			\#\#\# Round 2 \#\#\# \\
			User: The model trained with this set of hyperparameters has \{\{metric\}\} of \{\{value 1\}\} on the test dataset. Please provide a better set of hyperparameters. \\
			Assistant: \{\{Hyperparameter setting in JSON format\}\} \\
			\\
			\#\#\# Round 3 \#\#\# \\
			User: The model trained with this set of hyperparameters has \{\{metric\}\} of \{\{value 2\}\} on the test dataset. Please provide a better set of hyperparameters. \\
			Assistant: \{\{Hyperparameter setting in JSON format\}\} \\
			\hline
		\end{tabular}\label{tab:hpo_prompt_template}
	\end{table*}
	\begin{lstlisting}[basicstyle=\small,caption={JSON format of hyperparameter search space.},label={list:hpo_json_format}]
		{
			"iters": {
				"type": "number",
				"description": "The number of iterations of model training, an integer from 2000 to 7000."
			},
			"batch size": {
				"type": "number",
				"description": "Batch size during model training, an integer between 1 and 64."
			},
			"optimizer":{
				"type":"string",
				"enum": [
				"SGD",
				"Adam",
				"AdamW",
				"RMSprop"
				],
				"description":"Parameter optimizer for model training."
			},
			"learning rate":{
				"type":"number",
				"description":"Initial learning rate for model training."
			},
			"weight decay":{
				"type":"number",
				"description":"Weight decay value for model training."
			},
			"lr schedule": {
				"type": "string",
				"enum": [
				"MultiStepLR",
				"CosineAnnealingLR",
				"StepLR",
				"PolyLR"
				],
				"description":"Learning rate decay rules during model training."
			}
		}
	\end{lstlisting}
	
	\begin{lstlisting}[basicstyle=\small,caption={Example of data and model description.},label={list:hpo_example_datamodel}]
		{
			"data": {
				"num_classes": 4,
				"dataset": "ImageNet1k"
			},
			"model": {
				"name": "resnet34_8xb32_in1k",
				"params(M)": 2.18,
				"flops(G)": 3.68,
				"accuracy": 73.62
			}
		}
	\end{lstlisting}
	
	\begin{lstlisting}[basicstyle=\small,caption={Example of hyperparameter setting.},label={list:hpo_example_hp}]
		{
			"iters": 5000,
			"batch size": 8,
			"optimizer": "SGD",
			"learning rate": 6.2e-05,
			"weight decay": 0.02,
			"lr schedule": "StepLR"
		}
	\end{lstlisting}
	
	
	\section{Details of Experimental Setting and Results}
	\begin{table*}[htbp]
		\centering
		\caption{Numerical experimental results of hyperparameter optimization using random sampling. Evaluated by the mean and variance of the evaluation metrics.}
		\scalebox{0.98}{
			\begin{tabular}{|>{\centering\arraybackslash}p{0.06\textwidth}|>{\centering\arraybackslash}p{0.2\textwidth}|>{\centering\arraybackslash}p{0.2\textwidth}|>{\centering\arraybackslash}p{0.2\textwidth}|>{\centering\arraybackslash}p{0.2\textwidth}|}
				\hline
				\#Iter & Image classification & Object detection  & Semantic segmentation & Keypoint detection \\ \hline
				1      & 0.6704$\pm$0.0609    & 0.3087$\pm$0.0357 & 0.6144$\pm$0.0552     & 0.4570$\pm$0.0667  \\ \hline
				2      & 0.8156$\pm$0.0488    & 0.3866$\pm$0.0210 & 0.7460$\pm$0.0354     & 0.6103$\pm$0.0483  \\ \hline
				3      & 0.8869$\pm$0.0346    & 0.4133$\pm$0.0132 & 0.7960$\pm$0.0245     & 0.6683$\pm$0.0305  \\ \hline
				4      & 0.9200$\pm$0.0270    & 0.4249$\pm$0.0085 & 0.8196$\pm$0.0178     & 0.6932$\pm$0.0203  \\ \hline
				5      & 0.9418$\pm$0.0200    & 0.4310$\pm$0.0059 & 0.8326$\pm$0.0130     & 0.7060$\pm$0.0146  \\ \hline
				6      & 0.9539$\pm$0.0151    & 0.4345$\pm$0.0042 & 0.8398$\pm$0.0088     & 0.7135$\pm$0.0111  \\ \hline
				7      & 0.9614$\pm$0.0120    & 0.4368$\pm$0.0032 & 0.8440$\pm$0.0063     & 0.7183$\pm$0.0090  \\ \hline
				8      & 0.9665$\pm$0.0093    & 0.4384$\pm$0.0022 & 0.8466$\pm$0.0044     & 0.7226$\pm$0.0072  \\ \hline
				9      & 0.9702$\pm$0.0072    & 0.4395$\pm$0.0018 & 0.8482$\pm$0.0033     & 0.7252$\pm$0.0061  \\ \hline
				10     & 0.9723$\pm$0.0064    & 0.4404$\pm$0.0015 & 0.8493$\pm$0.0023     & 0.7277$\pm$0.0055  \\ \hline
			\end{tabular}
			\label{tab:hpo_random}
		}
	\end{table*}
	\subsection{Experimental Compute Resources and Hyperparameter Settings}
	\label{sec:experimental_setting}
	
	The fine-tuning of RU-LlaMA and HPO-LlaMA is executed on 8 NVIDIA Tesla A100 GPUs, each with 80G of GPU memory. The test of AutoMMLab is conducted on 8 NVIDIA Tesla V100 GPUs, each with 32G of GPU memory.
	When fine-tuning LlaMA2-7b to obtain RU-LlaMA and HPO-LlaMA, we used the AdamW optimizer to train 3 epochs. The initial learning rate was set to 1e-4, and decayed by 0.2 times after each epoch.
	The maximum text length is set to 4096 and the batch size is set to 4.
	The rank in LoRA was set to 8 and alpha to 32, fine-tuning the 'q\_proj' and 'v\_proj' layers. The mixed-precision approach is used for acceleration during training.
	When training multiple rounds of dialogue, only the output of the last round of dialogue is used as supervision information.
	
	\subsection{HPO Experimental Results of Random Sampling}
	Table~\ref{tab:hpo_random} presents the detailed quantitative results of hyperparameter optimization with the random sampling baseline.
	The evaluation metrics employed for image classification, object detection, semantic segmentation, and keypoint detection are accuracy, mean Average Precision (mAP), mean Intersection over Union (mIoU), and mAP, respectively.

	\subsection{Detailed HPO Experimental Results on LAMP}
	Table~\ref{tab:hpo_detail_cls}, Table~\ref{tab:hpo_detail_det}, Table~\ref{tab:hpo_detail_seg} and Table~\ref{tab:hpo_detail_pose} respectively list the experimental results of different models on the LAMP hyperparameter optimization dataset.
	It contains four computer vision tasks(image classification, semantic segmentation, object detection, keypoint detection), each task contains 20 requests equipped with corresponding train dataset, test dataset and model.
	The table shows the experimental results on the test dataset of models trained using hyperparameter configurations generated by different methods.
	
	\begin{table*}[htbp]
		\centering
		\caption{Detailed results of 20 classification requests. Evaluated by the accuracy on test dataset.}
		\resizebox{0.96\textwidth}{!}{
			\begin{tabular}{|c|c|c|c|c|c|c|c|c|c|c|}
				\hline
				\rowcolor{gray!50}
				\textbf{Model} & \textbf{1}  & \textbf{2}  & \textbf{3}  & \textbf{4}  & \textbf{5}  & \textbf{6}  & \textbf{7}  & \textbf{8}  & \textbf{9}  & \textbf{10} \\ \hline
				BayesianRF     & 0.333       & 0.760       & 0.750       & 0.960       & 0.795       & 0.920       & 0.760       & 0.333       & 0.500       & 0.640       \\ \hline
				BayesianGP     & 0.480       & 0.987       & 0.750       & 0.990       & 0.860       & 0.920       & 0.985       & 0.553       & 0.500       & 0.947       \\ \hline
				LLaMA2-7B-Chat & 0.965       & 0.333       & 0.947       & 0.750       & 0.960       & 0.985       & 0.993       & 0.990       & 0.653       & 0.500       \\ \hline
				PaLM2          & 0.975       & 0.993       & 0.993       & 0.750       & 0.990       & 1.00        & 1.00        & 0.995       & 0.993       & 0.950       \\ \hline
				GPT-3.5-turbo  & 0.955       & 0.720       & 0.973       & 0.750       & 0.980       & 0.970       & 0.980       & 0.985       & 0.333       & 0.500       \\ \hline
				GPT-4          & 0.965       & 0.333       & 0.967       & 0.750       & 0.960       & 0.980       & 0.973       & 0.975       & 0.660       & 0.500       \\ \hline
				HPO-LLaMA      & 0.975       & 0.993       & 0.973       & 0.995       & 0.980       & 0.995       & 1.00        & 0.995       & 1.00        & 0.940       \\ \hline
				\rowcolor{gray!50}
				\textbf{Model} & \textbf{11} & \textbf{12} & \textbf{13} & \textbf{14} & \textbf{15} & \textbf{16} & \textbf{17} & \textbf{18} & \textbf{19} & \textbf{20} \\ \hline
				BayesianRF     & 0.713       & 0.333       & 0.665       & 0.360       & 0.490       & 0.935       & 0.640       & 0.675       & 0.805       & 0           \\ \hline
				BayesianGP     & 0.753       & 0.333       & 0.940       & 0.870       & 0.945       & 0.935       & 0.940       & 0.710       & 0.825       & 0       \\ \hline
				LLaMA2-7B-Chat & 0.907       & 0.933       & 0.333       & 0.915       & 0.910       & 0.925       & 0.960       & 0.885       & 0.935       & 0.990       \\ \hline
				PaLM2          & 0.940       & 0.953       & 0.993       & 0.920       & 0.975       & 0.945       & 0.995       & 0.935       & 0.990       & 0.995       \\ \hline
				GPT-3.5-turbo  & 0.887       & 0.987       & 0.333       & 0.930       & 0.960       & 0.915       & 0.980       & 0.880       & 0.965       & 1.00        \\ \hline
				GPT-4          & 0.900       & 0.967       & 0.620       & 0.915       & 0.960       & 0.920       & 0.975       & 0.920       & 0.985       & 0.995       \\ \hline
				HPO-LLaMA      & 0.947       & 0.967       & 1.00        & 0.905       & 0.985       & 0.935       & 1.00        & 0.935       & 0.980       & 1.00        \\ \hline
			\end{tabular}\label{tab:hpo_detail_cls}}
		
		\vspace{0.5cm}
		\caption{Detailed results of 20 object detection requests. Evaluated by the mAP on test dataset.}
		\resizebox{0.96\textwidth}{!}{
			\begin{tabular}{|c|c|c|c|c|c|c|c|c|c|c|}
				\hline
				\rowcolor{gray!50}
				\textbf{Model} & \textbf{1}  & \textbf{2}  & \textbf{3}  & \textbf{4}  & \textbf{5}  & \textbf{6}  & \textbf{7}  & \textbf{8}  & \textbf{9}  & \textbf{10} \\ \hline
				BayesianRF     & 0.396       & 0           & 0.458       & 0.378       & 0.442       & 0           & 0.002       & 0.664       & 0.266       & 0.217       \\ \hline
				BayesianGP     & 0.179       & 0           & 0.466       & 0.377       & 0.217       & 0           & 0.003       & 0.698       & 0.269       & 0.240       \\ \hline
				LLaMA2-7B-Chat & 0           & 0           & 0.464       & 0           & 0           & 0.421       & 0.003       & 0           & 0.293       & 0.233       \\ \hline
				PaLM2          & 0.434       & 0.573       & 0.472       & 0.413       & 0.441       & 0.451       & 0.003       & 0.725       & 0.304       & 0.241       \\ \hline
				GPT-3.5-turbo  & 0.420       & 0.568       & 0.477       & 0.404       & 0.456       & 0.432       & 0.003       & 0.712       & 0.299       & 0.238       \\ \hline
				GPT-4          & 0.413       & 0.569       & 0.475       & 0.414       & 0.453       & 0.444       & 0.140       & 0.716       & 0.301       & 0.237       \\ \hline
				HPO-LLaMA      & 0.402       & 0.569       & 0.477       & 0.405       & 0.445       & 0.414       & 0.135       & 0.724       & 0.303       & 0.243       \\ \hline
				\rowcolor{gray!50}
				\textbf{Model} & \textbf{11} & \textbf{12} & \textbf{13} & \textbf{14} & \textbf{15} & \textbf{16} & \textbf{17} & \textbf{18} & \textbf{19} & \textbf{20} \\ \hline
				BayesianRF     & 0.483       & 0.519       & 0.206       & 0           & 0.342       & 0.558       & 0.029       & 0.456       & 0.275       & 0.123       \\ \hline
				BayesianGP     & 0.499       & 0.535       & 0.228       & 0.094       & 0.333       & 0.554       & 0.033       & 0.472       & 0.276       & 0.125       \\ \hline
				LLaMA2-7B-Chat & 0           & 0           & 0.220       & 0.100       & 0.371       & 0           & 0.030       & 0           & 0.301       & 0.126       \\ \hline
				PaLM2          & 0.510       & 0.540       & 0.225       & 0.104       & 0.378       & 0.590       & 0.031       & 0.477       & 0.305       & 0.128       \\ \hline
				GPT-3.5-turbo  & 0.511       & 0.539       & 0.224       & 0.104       & 0.376       & 0.586       & 0.031       & 0.468       & 0.302       & 0.128       \\ \hline
				GPT-4          & 0.507       & 0.535       & 0.217       & 0.359       & 0.373       & 0.590       & 0.636       & 0.475       & 0.296       & 0.537       \\ \hline
				HPO-LLaMA      & 0.507       & 0.532       & 0.230       & 0.372       & 0.377       & 0.586       & 0.660       & 0.485       & 0.301       & 0.539       \\ \hline
			\end{tabular}\label{tab:hpo_detail_det}}
		
		\vspace{0.5cm}
		\caption{Detailed results of 20 segmentation requests. Evaluated by the mIoU on test dataset.}
		\resizebox{0.96\textwidth}{!}{
			\begin{tabular}{|c|c|c|c|c|c|c|c|c|c|c|}
				\hline
				\rowcolor{gray!50}
				\textbf{Model} & \textbf{1}  & \textbf{2}  & \textbf{3}  & \textbf{4}  & \textbf{5}  & \textbf{6}  & \textbf{7}  & \textbf{8}  & \textbf{9}  & \textbf{10} \\ \hline
				BayesianRF     & 0.871       & 0.912       & 0.815       & 0.828       & 0.915       & 0.802       & 0.889       & 0.871       & 0.869       & 0.889       \\ \hline
				BayesianGP     & 0.861       & 0.923       & 0.832       & 0.828       & 0.902       & 0.802       & 0.889       & 0.866       & 0.864       & 0.891       \\ \hline
				LLaMA2-7B-Chat & 0           & 0           & 0           & 0           & 0           & 0           & 0           & 0           & 0.861       & 0           \\ \hline
				PaLM2          & 0.874       & 0.926       & 0.934       & 0.835       & 0.919       & 0.809       & 0.889       & 0.871       & 0.864       & 0.911       \\ \hline
				GPT-3.5-turbo  & 0.870       & 0.927       & 0.833       & 0.831       & 0.920       & 0.809       & 0.890       & 0.875       & 0.862       & 0.904       \\ \hline
				GPT-4-turbo    & 0           & 0.921       & 0.825       & 0.832       & 0.905       & 0.803       & 0.889       & 0.874       & 0.860       & 0.901       \\ \hline
				HPO-LLaMA      & 0.872       & 0.924       & 0.830       & 0.832       & 0.911       & 0.813       & 0.889       & 0.873       & 0.868       & 0.897       \\ \hline
				\rowcolor{gray!50}
				\textbf{Model} & \textbf{11} & \textbf{12} & \textbf{13} & \textbf{14} & \textbf{15} & \textbf{16} & \textbf{17} & \textbf{18} & \textbf{19} & \textbf{20} \\ \hline
				BayesianRF     & 0.874       & 0.765       & 0.807       & 0.861       & 0.828       & 0.895       & 0.792       & 0.820       & 0.825       & 0.809       \\ \hline
				BayesianGP     & 0.870       & 0.766       & 0.811       & 0.859       & 0.832       & 0.897       & 0.800       & 0.840       & 0.827       & 0.800       \\ \hline
				LLaMA2-7B-Chat & 0.868       & 0.748       & 0.763       & 0.864       & 0           & 0.900       & 0           & 0           & 0.822       & 0           \\ \hline
				PaLM2          & 0.872       & 0.757       & 0.627       & 0.864       & 0.830       & 0.906       & 0.796       & 0.851       & 0.824       & 0.842       \\ \hline
				GPT-3.5-turbo  & 0.873       & 0.761       & 0.797       & 0.862       & 0.827       & 0.897       & 0.800       & 0.847       & 0.827       & 0.833       \\ \hline
				GPT-4-turbo    & 0.870       & 0.746       & 0.793       & 0.861       & 0.825       & 0.899       & 0.799       & 0.821       & 0.825       & 0.814       \\ \hline
				HPO-LLaMA      & 0.871       & 0.763       & 0.809       & 0.867       & 0.828       & 0.913       & 0.801       & 0.849       & 0.828       & 0.839       \\ \hline
			\end{tabular}\label{tab:hpo_detail_seg}}
		
	\end{table*}
	
	\begin{table*}[htbp]
		\centering
		\caption{Detailed results of 20 keypoint detection requests. Evaluated by the mAP on test dataset.}
		\resizebox{0.96\textwidth}{!}{
			\begin{tabular}{|c|c|c|c|c|c|c|c|c|c|c|}
				\hline
				\rowcolor{gray!50}
				\textbf{Model} & \textbf{1}  & \textbf{2}  & \textbf{3}  & \textbf{4}  & \textbf{5}  & \textbf{6}  & \textbf{7}  & \textbf{8}  & \textbf{9}  & \textbf{10} \\ \hline
				BayesianRF     & 0.457       & 0           & 0.208       & 0           & 0.092       & 0           & 0           & 0           & 0           & 0       \\ \hline
				BayesianGP     & 0           & 0           & 0.078       & 0.173       & 0           & 0.118       & 0.359       & 0           & 0           & 0       \\ \hline
				LLaMA2-7B-Chat & 0           & 0           & 0           & 0           & 0           & 0           & 0           & 0           & 0           & 0           \\ \hline
				PaLM2          & 0.883       & 0.716       & 0.705       & 0.686       & 0.780       & 0.566       & 0.733       & 0.706       & 0.781       & 0.698       \\ \hline
				GPT-3.5-turbo  & 0.515       & 0.285       & 0.192       & 0.131       & 0.164       & 0.392       & 0.329       & 0.147       & 0.148       & 0.170       \\ \hline
				GPT-4-turbo    & 0.218       & 0.002       & 0.002       & 0.000       & 0.058       & 0.566       & 0.003       & 0.003       & 0.039       & 0.262       \\ \hline
				HPO-LLaMA      & 0.863       & 0.741       & 0.711       & 0.642       & 0.743       & 0.755       & 0.717       & 0.686       & 0.768       & 0.691       \\ \hline
				\rowcolor{gray!50}
				\textbf{Model} & \textbf{11} & \textbf{12} & \textbf{13} & \textbf{14} & \textbf{15} & \textbf{16} & \textbf{17} & \textbf{18} & \textbf{19} & \textbf{20} \\ \hline
				BayesianRF     & 0           & 0.401       & 0           & 0.135       & 0           & 0           & 0           & 0.004       & 0.065       & 0.020       \\ \hline
				BayesianGP     & 0           & 0.249       & 0           & 0           & 0           & 0.044       & 0           & 0.558       & 0           & 0.039       \\ \hline
				LLaMA2-7B-Chat & 0           & 0           & 0           & 0           & 0           & 0           & 0           & 0           & 0           & 0           \\ \hline
				PaLM2          & 0.788       & 0.529       & 0.660       & 0.758       & 0.782       & 0.734       & 0.790       & 0.674       & 0.690       & 0.726       \\ \hline
				GPT-3.5-turbo  & 0.398       & 0.529       & 0.102       & 0.305       & 0.036       & 0.028       & 0.046       & 0.077       & 0.078       & 0.004       \\ \hline
				GPT-4-turbo    & 0.016       & 0.172       & 0.074       & 0.419       & 0.000       & 0.002       & 0.000       & 0.042       & 0.040       & 0.000       \\ \hline
				HPO-LLaMA      & 0.764       & 0.786       & 0.663       & 0.698       & 0.764       & 0.719       & 0.776       & 0.677       & 0.703       & 0.688       \\ \hline
			\end{tabular}\label{tab:hpo_detail_pose}}
	\end{table*}
	
	\section{Limitations}
	\label{sec:limitations}
	AutoMMLab has presented a general-purpose LLM-empowered AutoML system that follows user’s language instructions to automate the whole model production workflow for computer vision tasks. However there still remain some limitations and improvement spaces: \textbf{1) Request Understanding}: Whether the generated models can meet the user's requirements perfectly depends largely on the accuracy of understanding the user's requests. Understanding the multimodal information input by the users and interacting with the users in a multimodal form can more accurately understand the user's requests. \textbf{2) Data Selection}: Our method builds a dataset zoo to find images related to user's requests and build training datasets. Model generation will fail if no images related to the user's requests are found in the dataset zoo. Building a more powerful database and generating training data by generating models can effectively improve the success rate of model generation. \textbf{3) Model Selection}: The models supported by our method are limited to those already available in OpenMMLab. Personalized customized models according to user descriptions can meet more diverse model requirements of users. \textbf{4) LLM Hallucination}: Although LLMs have the strong generative ability, they possibly still failure to follow instructions present hallucination. How to mitigate hallucination in LLMs and reduce the effects of hallucination should be considered when designing the system.
	\section{Broad impact and ethics statement}
	\label{sec:impacts}
	AutoMMLab is expected to have extensive benefits, facilitating AI-driven innovation and empowering individuals with limited AI expertise to leverage AI capabilities. 
	Additionally, AutoMMLab offers valuable resources for computer vision researchers, serving as a reference for task baselines and providing a platform for evaluating and developing end-to-end prompt-based model training. We plan to release AutoMMLab as an open-source project, encouraging community contributions.
	
	The accessibility of any powerful technology to the general public raises ethical considerations~\cite{widder2022limits}, particularly concerning the potential misuse by malicious actors.
	The deployment of AutoMMLab poses a risk of generating models that may produce toxic or offensive content when exposed to harmful input. To address these ethical concern, our design emphasizes transparency throughout the entire process and incorporates rigorous security measures in the construction of dataset zoo and model zoo, ensuring the reliability of the generated models.
	
	\section{Comparisons with other AutoML systems}
	AutoMMLab is a research project, whose goal is not to create a commercial system, but to showcase how LLM-empowered AutoML system can achieve fully end-to-end prompt-based model training, which is expected to have broad impact for the community. 
	In Table~\ref{tab:compare_automl}, we compare with several AutoML systems for reference. As far as we know, we are the \emph{first} to build a request-to-model AutoML system for computer vision tasks.
	Previous AutoML systems are unable to perform our newly proposed ``end-to-end prompt-based model training'' task.
	\begin{figure}[htbp]
		\centering
		\caption{Comparisons with other AutoML systems in the following dimensions: Request Understanding (RU), Programming-Free (PF), Data Preparation (DP), Model Preparation (MP), Hyperparameter Optimizer (HPO), Supporting Computer Vision tasks (SCV). \cmark ~means automatic, \cxmark ~means requiring manual operation, \xmark ~means unsupported.}
		\resizebox{0.49\textwidth}{!}{
			\begin{tabular}{lcccccc}
				\toprule
				AutoML Systems                              & RU & PF & DP & MP & HPO & SCV \\ \hline
				AutoKeras~\cite{jin2023autokeras}           & \xmark & \xmark  & \xmark  & \cxmark  & \cmark   & \xmark   \\
				Amazon SageMaker~\cite{das2020amazon}       & \xmark & \xmark & \cxmark  & \cxmark  & \cmark   & \cmark   \\
				Google AutoML~\cite{bisong2019google}       & \xmark & \cmark & \cxmark  & \cmark  & \cmark   & \cmark   \\
				Prompt2Model~\cite{prompt2model}            & \cxmark & \cmark & \cmark  & \cmark  & \xmark   & \xmark\\
				\rowcolor{ourscolor}
				AutoMMLab                                   & \cmark  & \cmark  & \cmark  &  \cmark  &  \cmark &  \cmark   \\
				\bottomrule
			\end{tabular}
		}
		\label{tab:compare_automl}
	\end{figure}
	\section{Example of End-to-end Model Production}
\begin{figure*}[thbp]
		\centering
		\includegraphics[height=0.98\textheight]{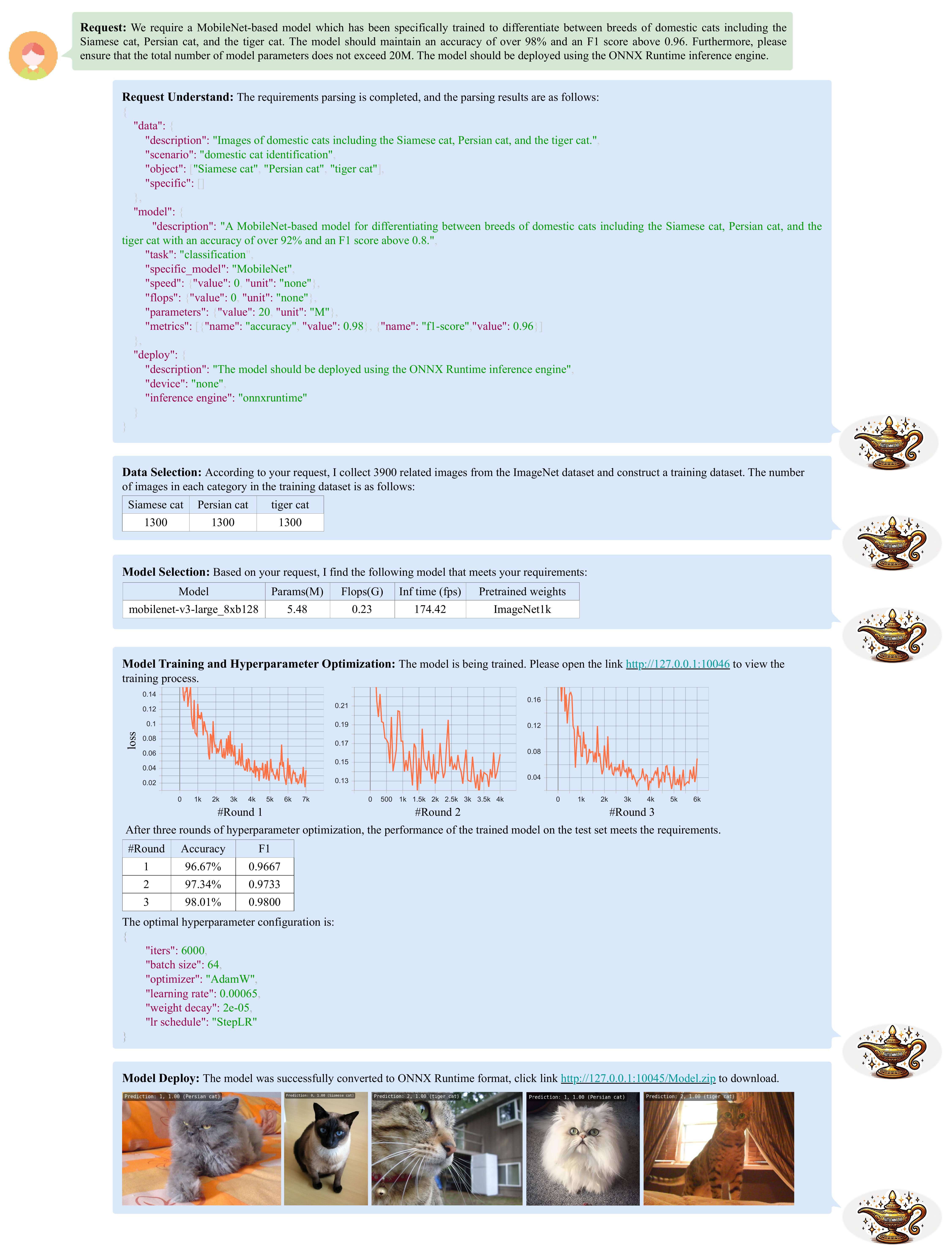}
		\caption{Example of end-to-end model production workflow.} 
		\label{fig:e2e_example}
	\end{figure*}

	Figure~\ref{fig:e2e_example} exhibits an example of end-to-end model production workflow.
	Upon receiving a user request,  our AutoMMLab automatically executes request understanding, data selection, model selection, model training, hyperparameter optimization, and model deployment, generating a model that can be directly deployed on the specific device.
	
\end{document}